\documentclass[journal,comsoc]{IEEEtran}
\usepackage{longtable}
\usepackage{times}
\usepackage{epsfig}
\usepackage{graphicx}
\usepackage{amsmath}
\usepackage{amssymb}
\usepackage{epsfig}
\usepackage{makecell,multirow,diagbox}
\usepackage[normalem]{ulem}
\usepackage{ulem}
\usepackage{comment}
\usepackage[colorlinks,linkcolor=red]{hyperref}
\usepackage{footnote}
\usepackage{threeparttable}
\usepackage{subfigure}

\def\ie{\emph{i.e.}}
\usepackage{xcolor,soul,framed} 
\usepackage{array}
\usepackage{mdwmath}
\usepackage{mdwtab}
\usepackage{booktabs}
\usepackage[top=2cm, bottom=2cm, left=2cm, right=2cm]{geometry}
\usepackage{algorithm}
\usepackage{algpseudocode}

\newcommand{\tabincell}[2]{\begin{tabular}{@{}#1@{}}#2\end{tabular}}
\ifCLASSINFOpdf
\else
\fi
\usepackage{amsmath}
\usepackage[cmintegrals]{newtxmath}
\makeatletter
\renewcommand{\@thesubfigure}{\hskip\subfiglabelskip}
 \makeatother

\begin{document}
%
\title{Transform consistency for learning with noisy labels}
%
%
%

\author{Rumeng~Yi,
        Yaping~Huang
\thanks{Rumeng Yi, Yaping Huang are with the Beijing Key Laboratory of Traffic Data Analysis and Mining, Beijing Jiaotong University, Beijing 100044, China (e-mail: 19112038@bjtu.edu.cn; yphuang@bjtu.edu.cn).}}

%
%

\markboth{}%
{ Transform consistency for learning with noisy labels}
%



\maketitle

\begin{abstract}
It is crucial to distinguish mislabeled samples for dealing with noisy labels. Previous methods such as ``Co-teaching'' and ``JoCoR'' introduce two different networks to select clean samples out of the noisy ones and only use these clean ones to train the deep models. Different from these methods which require to train two networks simultaneously, we propose a simple and effective method to identify clean samples only using one single network. We discover that the clean samples prefer to reach consistent predictions for the original images and the transformed images while noisy samples usually suffer from inconsistent predictions. Motivated by this observation, we introduce to constrain the transform consistency between the original images and the transformed images for network training, and then select small-loss samples to update the parameters of the network. Furthermore, in order to mitigate the negative influence of noisy labels, we design a classification loss by using the off-line hard labels and on-line soft labels to provide more reliable supervisions for training a robust model. We conduct comprehensive experiments on CIFAR-10, CIFAR-100 and Clothing1M datasets. Compared with the baselines, we achieve the state-of-the-art performance. Especially, in most cases, our proposed method outperforms the baselines by a large margin.
\end{abstract}


%
\IEEEpeerreviewmaketitle

\section{Introduction}
%
%
%
%
\IEEEPARstart{D}{eep} Neural Networks (DNNs) achieve a remarkable success on many computer vision tasks due to the large-scale datasets with reliable and clean annotations~\cite{chen2018encoder}~\cite{girshick2015fast}~\cite{he2016deep}. However, collecting such datasets with precise annotations is expensive and time-consuming. There are two alternative solutions to alleviate this issue: crowd-sourcing from non-experts and online queries by search engines. Unfortunately, the obtained annotations inevitably contain noisy labels. As DNNs have the capability to memorize all training samples, they will eventually overfit the noisy labels, leading to poor generalization performance~\cite{tanaka2018joint}~\cite{zhang2016understanding}.

\begin{figure}[t]
\centering
\subfigure[(a) Symmetric 50\%]{
\begin{minipage}[t]{0.45\linewidth}
\centering
\includegraphics[width=1.5in]{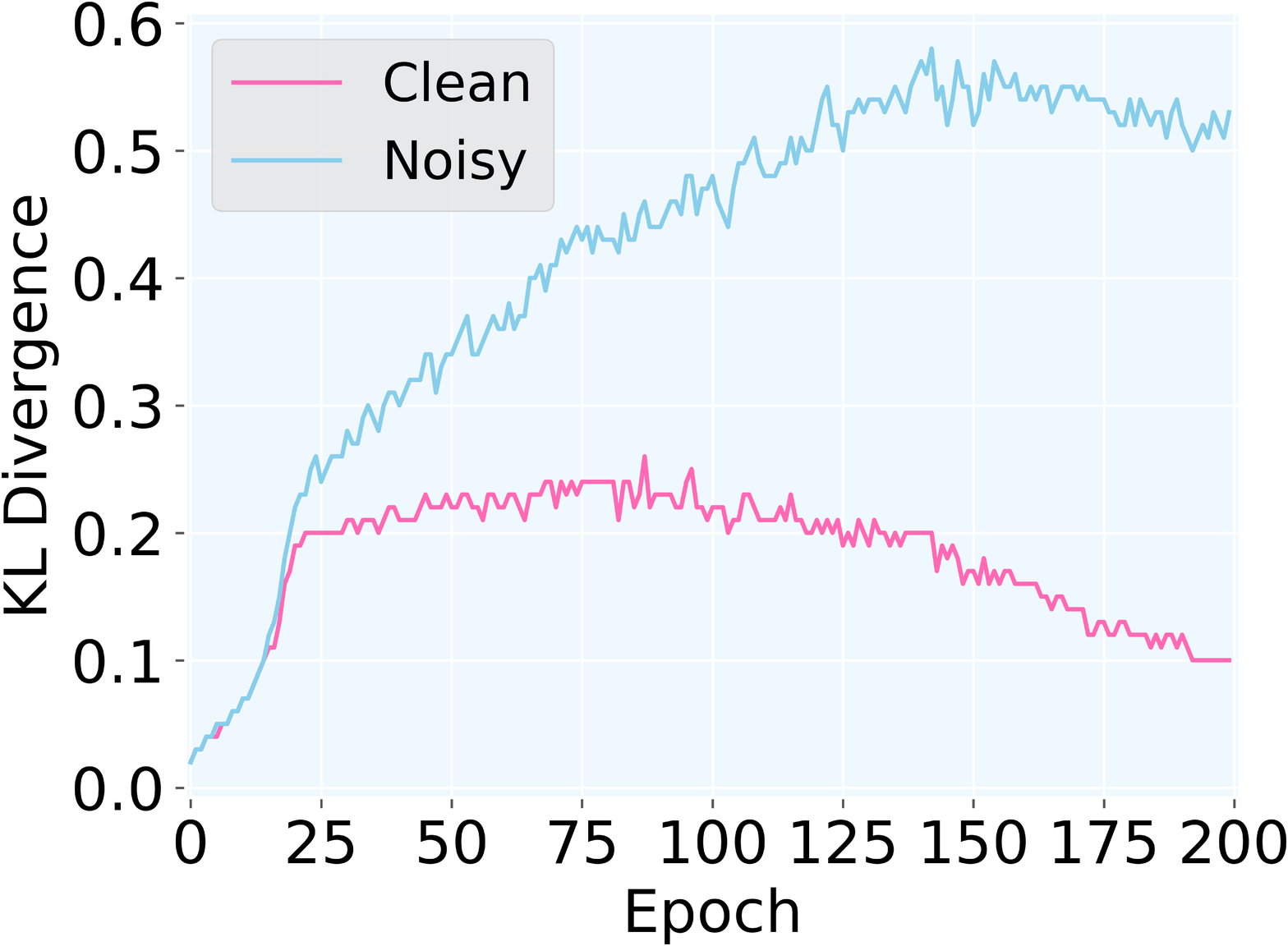}
\end{minipage}
}
\subfigure[(b) Asymmetric 40\%]{
\begin{minipage}[t]{0.45\linewidth}
\centering
\includegraphics[width=1.5in]{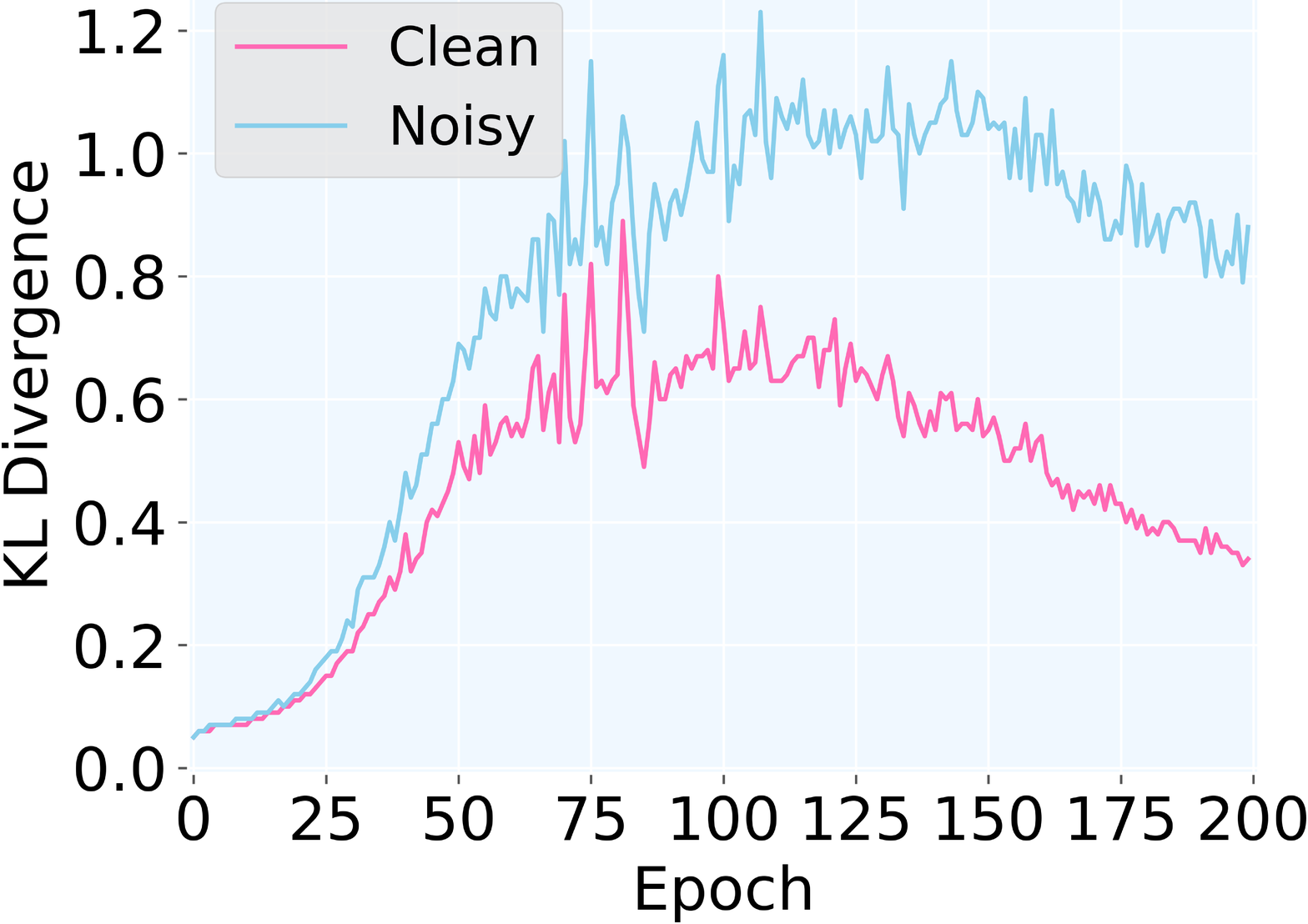}
\end{minipage}
}
\caption{KL Divergence between two inputs (original and horizontally flip images) on CIFAR-10 under (a) symmetric 50\% and (b) asymmetric 40\% label noise for clean and noisy samples.}
\label{KL Divergence}
\end{figure}

In order to reduce the effect of noisy labels, an efficient strategy is to select clean samples out of the noisy ones and then only use these clean ones to update the parameters of the network. Following this research line, many studies have shown impressive progresses, where Decoupling~\cite{malach2017decoupling}, Co-teaching~\cite{han2018co} and Co-teaching+~\cite{yu2019does} are three representative methods. All these methods train two networks simultaneously, but the difference is that Decoupling and Co-teaching+ select training samples depending on the disagreement between two different networks, while in Co-teaching, each network views its small-loss instances as clean ones, and teaches them to its peer network. Recently, the state-of-the-art method named JoCoR~\cite{wei2020combating} achieves the excellent performance. JoCoR applies an agreement maximization principle that trains two networks with a joint loss (including the conventional supervised loss and the co-regularization loss) and use the joint loss to select small-loss instances. 

In this paper, different from the above methods which require to train two networks simultaneously, we propose a simple and effective method to distinguish clean samples only using one single network. We find that the prediction consistency under different image transforms (such as scaling, rotation, flipping) in one network is beneficial to select clean samples. As shown in Figure~\ref{KL Divergence}, we feed the original and transformed (horizontally flip) images into one single network, and observe the Kullback-Leibler (KL) Divergence of clean and noisy samples on CIFAR-10 dataset with 50\% symmetric noise and 40\% asymmetric noise. From Figure~\ref{KL Divergence}, we can see that the KL Divergence of clean samples are much smaller than noisy
samples under two different noise levels. It suggests that the network can reach an agreement on the original and transformed images for clean samples during training. While for noisy samples, the KL Divergence remains a larger value. This observation motivates us to distinguish clean and noisy samples by considering the KL Divergence between the original images and the transformed images.

Based on the above observation, we propose a novel approach to identify mislabeled samples. Specifically, we follow the agreement maximization principle to train a network with a joint loss, including the classification losses and the KL loss between two inputs (the original images and transformed images). Moreover, we further adapt a self-ensembling method to construct a teacher model, the predictions of the teacher model are used as the on-line soft labels. Then we combine off-line hard labels with on-line soft labels to provide robust supervisory information for training and further alleviate the noise effects.

Intuitively, our proposed method can be regarded as adding some perturbations on the training samples. Since we observe that the perturbation can have stronger effect on the noisy samples than that on clean samples, making clean and noisy samples being distinguished more easily. Furthermore, our method only requires to train one single network, which is more suitable for real applications.

In summary, our contribution in this paper are three-fold:
\begin{itemize}
  \item We propose a simple and effective approach to learn from noisy labels, which introduces the prediction consistency of the images under different spatial transforms (such as scaling, rotation, flipping) in one single network to select clean samples efficiently.
   
  \item We design a classification loss by using the off-line hard labels and on-line soft labels to provide  reliable supervisions for network training.
  
  \item We conduct comprehensive experiments on CIFAR-10, CIFAR-100 and Clothing1M datasets, and achieve the state-of-the-art performance. In most cases, such as symmetry-50\%, symmetry-80\% and asymmetry-40\% on CIFAR-10, and symmetry-20\%, symmetry-50\% and symmetry-80\% on CIFAR-100, the performance of our method exceeds the second best method by up to 3.70\%-12.78\%.
  
\end{itemize}

\section{Related Works}
\subsection{Learning with Noisy Labels}
The techniques to alleviate the effect of noisy labels can be divided into two categories: (1) detecting noisy labels and then cleansing potential noisy labels or reduce their impacts. (2) directly training noise-robust models with noisy labels.

One solution of detecting and cleansing noisy labels is to select clean samples out of the noisy ones, and then use them to update the network. For example, Decoupling~\cite{malach2017decoupling}, Co-teaching~\cite{han2018co},  Co-teaching+~\cite{yu2019does} and the recent state-of-the-art JoCoR~\cite{wei2020combating} train two networks by introducing ``agreement'' or ``disagreement'' strategies to select clean samples for training. O2U-Net~\cite{huang2019o2u}
adjusts the hyperparameters of the network and ranks the normalized average loss of every sample to detect the noisy samples. MentorNet~\cite{jiang2018mentornet} learns a data-driven curriculum to supervise the training of a student network. The second attempt is based on re-weighting or relabeling methods. Han et al.~\cite{han2019deep} propose a deep self-learning framework to relabel the noisy samples by extracting multiple prototypes for one category. Ren et al.~\cite{ren2018learning} weights each sample in the loss function based on the gradient directions compared to those on a clean set. Arazo et al.~\cite{arazo2019unsupervised} model the per-sample loss distribution with a beta mixture model and correct the loss by relying on the network prediction. In a similar spirit, Li et al.~\cite{li2020dividemix} use two networks to perform sample selection via a gaussian mixture model and further apply the semi-supervised learning technique, MixMatch~\cite{berthelot2019mixmatch}, to handle noisy labels.

For the second type, Xiao et al.~\cite{xiao2015learning} model the relationships between images, class labels and label noises with a probabilistic graphical model and further integrate it into an end-to-end deep learning model. Guo et al.~\cite{guo2018curriculumnet} propose a CurriculumNet, where training data are divided into several subsets by ranking their complexity via distribution density, and then the subsets are formed as a curriculum to teach the model in understanding noisy labels gradually. Li et al.~\cite{li2019learning} propose a meta-learning based method to avoid the model overfitting to the specific noise by generating synthetic noisy labels.

Motivated by the above studies, we propose a simple and effective method to select clean samples out of the noisy ones by observing the category distribution and visual attention map between the original images and the transformed images in one single network. This achieves an aggregated inference which combines the predictions from different spatial transforms to improve the classification accuracy.

\subsection{Image Transformation}
Human visual perception shows good consistency under certain spatial transforms, such as scaling, rotation and flipping, which has motivated the data augmentation strategy widely used in DNNs. \cite{gidaris2018unsupervised} learns image representations by training convolutional networks to recognize the geometric transformation that is applied to the input image. \cite{guo2019visual} proposes a two-branch network with original images and its transformed images as inputs and introduces a new attention consistency loss that measures the attention heatmap consistency between two branches and achieves a new state-of-the-art for multi-label image classification.
~\cite{lee20_sla} augments original labels via self-supervision of input transformation to learn a single unified task with respect to the joint distribution of the original and self-supervised labels.

In this paper, we propose the first usage of the transform consistency  for noisy-labeled image classification. Specifically, we regard the transformation operation performed on the images as the added disturbance. For clean samples, the network's predictions are not affected by the added disturbance, resulting in a consistent prediction between two inputs, while for noisy samples, the network's predictions may be inconsistent or oscillate strongly, which allow us to distinguish the clean samples out of the noisy ones.

\begin{figure*}[t]
\begin{center}
\setlength{\abovecaptionskip}{0pt}
\includegraphics[width=1\linewidth]{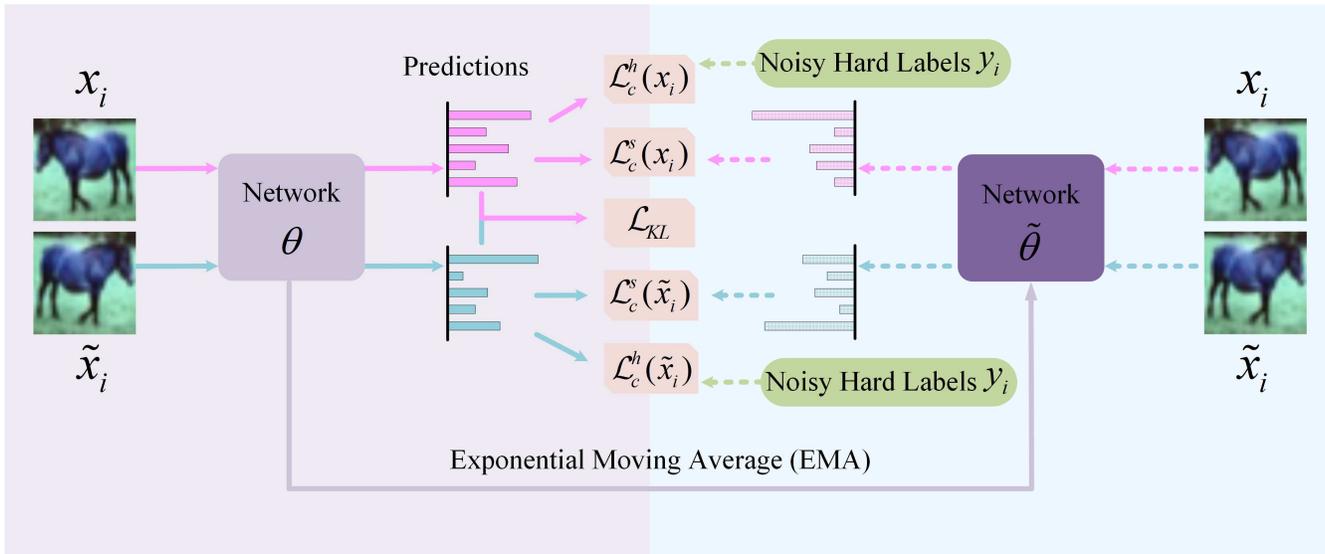}
\caption{Overview of the proposed approach. Specifically, an original image and its transformed image are both fed into one single network, and then apply a joint loss, \ie, two classification loss ($\mathcal{L}_{c}^{h}$ and $\mathcal{L}_{c}^{s}$) and one KL loss (${{\mathcal{L}}_{KL}}$) to train the network. Finally we select the samples with small joint loss to update the model’s parameters ($\theta$). Besides, we use the exponential moving average (EMA) to construct a teacher model, the prediction of the two inputs are used as the on-line soft labels for optimizing $\theta$.
}
\label{framework}
\end{center}
\end{figure*}

\subsection{Self-ensemble Learning} Self-ensemble has been widely studied in semi-supervised learning. These methods form a consensus prediction of the unknown labels using the outputs of the network in training on different epochs. For example, temporal ensembling~\cite{laine2016temporal} maintains an exponential moving average (EMA) of label predictions on each training example, and penalizes predictions that are inconsistent with this target. Mean teacher~\cite{tarvainen2017mean} averages model weights over training steps to form a target-generating teacher model. Recently, the self-ensembling strategy is used for noisy-labeled image classification.
MLNT~\cite{li2019learning} designs a noise-tolerant algorithm which constructs a teacher model to give more reliable predictions unaffected by the synthetic noise. SELF~\cite{nguyen2019self} attempts to identify clean labels progressively using self-forming ensembles of models and predictions.

To effectively mitigate the negative influence of noisy labels, in this paper, we introduce the self-ensembling method to construct a teacher model by using the exponential moving average (EMA) of the model snapshots in each training iterations. Then the predictions of the teacher model are used as the on-line soft labels, which provide a more stable supervisory signal than noisy hard labels to guide the model's training. Note that this teacher model does not require additional training.

In this section, we give the details of our proposed method. Our approach aims at identifying potentially noisy labels during training and keeping the network from receiving supervision of the clean labels. The overall framework is illustrated in Figure~\ref{framework}. Specifically, an original image and its transformed image are both fed into one single network, and then we apply a joint loss, \ie, two classification losses and one KL loss, to train the network. Based on our observed principle that the model can keep consistent predictions on two inputs of clean samples and inconsistent predictions on noisy labels, we select the samples with small joint loss to update the model's parameters. In order to further alleviate the negative influence of label noise, apart from the off-line hard labels, our framework further incorporates on-line soft pseudo labels by using the self-ensembling strategy into the training process.

\textbf{Problem statement.}
Image classification can be formulated with a training set $\mathcal{D}=\{({{x}_{1}},{{y}_{1}}),...,({{x}_{n}},{{y}_{n}})\}$, where ${{x}_{i}}$ is an image and ${{y}_{i}}\in {{\{0,1\}}^{C}}$ is the one-hot label over $C$ classes which may contain noise. Let $f({{x}_{i}};\theta )$ denote the model's output softmax probability parameterized by $\theta $.

\textbf{Classification loss.}
In our framework, the two inputs of the network are ${{x}_{i}}$ and ${{\tilde{x}}_{i}}$, respectively, where ${{\tilde{x}}_{i}}=t({{x}_{i}})$ and $t$ represents the transformation operation such as scaling, rotation and flipping. Their label confidences can be predicted as $f({{x}_{i}};\theta )$ and $f({{\tilde{x}}_{i}};\theta )$. In general, the objective function is the empirical risk of cross-entropy loss, which is formulated by:
\begin{equation}
{{\mathcal{L}}_{c}^{h}}({{x}_{i}})=-\frac{1}{N}\sum\limits_{i=1}^{N}{{{y}_{i}}\cdot \log (f({{x}_{i}};\theta ))}
\end{equation}
\begin{equation}
    {{\mathcal{L}}_{c}^{h}}({{\tilde{x}}_{i}})=-\frac{1}{N}\sum\limits_{i=1}^{N}{{{y}_{i}}\cdot \log (f({{{\tilde{x}}}_{i}};\theta ))}
\end{equation}
where $N$ is the total number of samples. However, since ${{y}_{i}}$ contains noise, the model would overfit the noisy labels and result a poor classification performance.

To address this problem, we construct a teacher model parameterized by $\tilde{\theta }$ following the self-ensembling method to generate reliable soft pseudo labels for supervising the model's training. Specifically, at current iteration $T$, we update ${{\tilde{\theta }}^{(T)}}$ with
\begin{equation}
{{\tilde{\theta }}^{(T)}}=\gamma {{\tilde{\theta }}^{(T-1)}}+(1-\gamma )\theta
\label{ensembling}
\end{equation}
where $\gamma $ is a smoothing coefficient hyperparameter within the range [0,1), and ${{\tilde{\theta }}^{(T-1)}}$ indicates the parameter of the teacher model in the previous iteration $(T-1)$. Then the soft classification loss for optimizing $\theta $ with the soft pseudo labels generated from the teacher model can be formulated as:
\begin{equation}
    \mathcal{L}_{c}^{s}({{x}_{i}})=-\frac{1}{N}\sum\limits_{i=1}^{N}{f({{x}_{i}};\tilde{\theta })}\cdot \log (f({{x}_{i}};\theta ))
\end{equation}
\begin{equation}
    \mathcal{L}_{c}^{s}({{\tilde{x}}_{i}})=-\frac{1}{N}\sum\limits_{i=1}^{N}{f({{{\tilde{x}}}_{i}};\tilde{\theta })}\cdot \log (f({{\tilde{x}}_{i}};\theta ))
\end{equation}
Therefore, the classification loss of ${{x}_{i}}$ and ${{\tilde{x}}_{i}}$ are given by:
\begin{equation}
    {{\mathcal{L}}_{c}}({{x}_{i}})=\mathcal{L}_{c}^{h}({{x}_{i}})+\mathcal{L}_{c}^{s}({{x}_{i}})
\label{soft_loss1}
\end{equation}

\begin{equation}
    {{\mathcal{L}}_{c}}({{\tilde{x}}_{i}})=\mathcal{L}_{c}^{h}({{\tilde{x}}_{i}})+\mathcal{L}_{c}^{s}({{\tilde{x}}_{i}})
\label{soft_loss2}
\end{equation}

\textbf{KL loss.}
Based on the observation that the model can make consistent predictions on the original and transformed images of clean samples, and inconsistent predictions on most noisy samples, we apply the Kullback-Leibler (KL) Divergence as KL loss to measure the probability distributions of the network for two inputs. In practice, the KL loss can be expressed as
\begin{equation}
{{\mathcal{L}}_{KL}}={{D}_{KL}}({{p}_{x}}\parallel {{p}_{{\tilde{x}}}})+{{D}_{KL}}({{p}_{{\tilde{x}}}}\parallel {{p}_{x}})
\end{equation}
where
\begin{equation}
    {{D}_{KL}}({{p}_{x}}\parallel {{p}_{{\tilde{x}}}})=\sum\nolimits_{i=1}^{N}{\sum\nolimits_{j=1}^{C}{{{p}^{j}}({{x}_{i}})\log \frac{{{p}^{j}}({{x}_{i}})}{{{p}^{j}}({{{\tilde{x}}}_{i}})}}}
\end{equation}
\begin{equation}
    {{D}_{KL}}({{p}_{{\tilde{x}}}}\parallel {{p}_{x}})=\sum\nolimits_{i=1}^{N}{\sum\nolimits_{j=1}^{C}{{{p}^{j}}({{{\tilde{x}}}_{i}})\log \frac{{{p}^{j}}({{{\tilde{x}}}_{i}})}{{{p}^{j}}({{x}_{i}})}}}.
\end{equation}
Here $C$ is the number of categories.  ${{p}^{j}}({{x}_{i}})$ and ${{p}^{j}}({{\tilde{x}}_{i}})$ represent the predicted probability distributions, \ie, ${{p}}({{x}_{i}})={{f}}({{x}_{i}};\theta )$ and ${{p}}({{\tilde{x}}_{i}})={{f}}({{\tilde{x}}_{i}};\theta )$, repectively.

\textbf{Overall loss.}
We integrate the above-mentioned losses. The total loss can be formulated as:
\begin{equation}
    {{\mathcal{L}}_{total}}=(1-\lambda )({{\mathcal{L}}_{c}}({{x}_{i}})+{{\mathcal{L}}_{c}}({{\tilde{x}}_{i}}))+\lambda {{\mathcal{L}}_{KL}}
\label{total loss}
\end{equation}
where $\lambda $ is the parameter weighting the classification loss and KL loss.

\textbf{Small-loss selection.}
We use the total loss to select the small-loss instances. Specifically, the network feeds forward a mini-batch of data $\bar{\mathcal{D}}=\{({{x}_{1}},{{y}_{1}}),({{x}_{2}},{{y}_{2}}),...,({{x}_{B}},{{y}_{B}})\}$ from $\mathcal{D}$ first, then we select a small proportion of instances ${{\bar{\mathcal{D}}}^{'}}$ in $\bar{\mathcal{D}}$ which have small total loss:
\begin{equation}
{{\bar{\mathcal{D}}}^{'}}=\arg {{\min }_{{{\mathcal{D}}^{'}}:|{{\mathcal{D}}^{'}}|\ge R(T)|\bar{\mathcal{D}}|}}{{\mathcal{L}}_{total}}({{\mathcal{D}}^{'}})
\label{small-dataset}
\end{equation}
where $R(T)$ is a parameter that controls how many small-loss instances should be selected. For $R(T)$, we follow the same update strategy as \cite{han2018co}~\cite{yu2019does}~\cite{wei2020combating}. Specifically, since the DNNs start to learn from clean samples in initial phases and gradually adapt to noisy ones during training, we apply a large $R(T)$ to use more small-loss instances to train the model in the beginning. With the increase of epoch, as the DNNs will overfit the noisy data gradually, we should update the $R(T)$ with a smaller value to keep less small-loss instances selected in each mini batch. The update strategy for $R(T)$ is given by:
\begin{equation}
    R(T)=1-\min \{\frac{T}{{{T}_{k}}}\tau ,\tau \}
    \label{update rt}
\end{equation}
where $T$ is the current iteration, ${{T}_{k}}$ represents how many epochs for linear drop rate, and $\tau $ is the actual noise rate in the whole dataset which is closely related to the noise ratio in the noisy classes. If $\tau $ is not known in advanced, $\tau $ can be inferred using validation set~\cite{liu2015classification}~\cite{yu2018efficient}.

  \begin{algorithm}[t]  
  \caption{}  
  \label{alg:Framwork}  
  \begin{algorithmic}[1]  
    \Require  
      $\theta$ and $\tilde{\theta }$, training set $\mathcal{D}$, learning rate $\eta$, fixed $\tau$, epoch ${{T}_{k}}$ and ${{T}_{max}}$, iteration ${{N}_{max}}$, ensembling momentum $\gamma$ ;
    \For{$T$=1,...,${{T}_{max}}$}
    \State \textbf{Shuffle} training set $\mathcal{D}$;
        \For{$N$=1,...,${{N}_{max}}$}
        \State \textbf{Fetch} mini-batch ${\bar{\mathcal{D}}}$ from $\mathcal{D}$;
        \State \textbf{Calculate} the total loss ${{\mathcal{L}}_{total}}$ by Eq. (\ref{total loss});
        \State \textbf{Obtain} small-loss sets ${{{\bar{\mathcal{D}}}}^{'}}$ by Eq. (\ref{small-dataset}) from ${\bar{\mathcal{D}}}$;
        \State \textbf{Obtain} $\mathcal{L}$ by Eq. (\ref{small-loss}) on ${{{\bar{\mathcal{D}}}}^{'}}$;
        \State \textbf{Update} ${\theta \text{ = }\theta \text{ - }\eta \nabla \mathcal{L}}$;
        \EndFor
    \State \textbf{Update} ${{\tilde{\theta }}^{(T)}}=\text{ }\gamma {{\tilde{\theta }}^{(T-1)}}\text{ }+\text{ }(1-\gamma )\theta $ by Eq. (\ref{ensembling});
    \State \textbf{Update} ${R(T)=1-\min \{\frac{T}{{{T}_{k}}}\tau ,\tau \}}$ by Eq. (\ref{update rt});
    \EndFor
\Ensure  
$\theta$ and $\tilde{\theta }$.
  \end{algorithmic}  
\end{algorithm}  

Finally, we calculate the average loss of the selected instances ${{\bar{\mathcal{D}}}^{'}}$ and back-propagate them to update the network's parameters:
\begin{equation}
    \mathcal{L}=\frac{1}{\left| {{{\bar{\mathcal{D}}}}^{'}} \right|}\sum\nolimits_{x\in {{{\bar{\mathcal{D}}}}^{'}}}{{{\mathcal{L}}_{total}}(x)}
    \label{small-loss}
\end{equation}

Algorithm 1 delineates the overall training process.

\section{Experiments}
\subsection{Datasets and implementation details}
\textbf{Datasets.}
We extensively evaluate our approach on CIFAR-10, CIFAR-100~\cite{krizhevsky2009learning} and Clothing1M~\cite{xiao2015learning} datasets. Both CIFAR-10 and CIFAR-100 contain 50K training images and 10K test images of size 32 $\times$ 32, which involve 10 classes and 100 classes, respectively. Clothing1M contains 1 million images, which consists of 50k training images, 14k validation images and 10k test images. Human annotators are asked to mark a set of 25k labels as a clean set, but we do not use these in our experiments. Note that the overall label accuracy of Clothing1M is 61.54\%.

\textbf{Implementation.} For all experiments in our paper, we follow the same setting as JoCoR~\cite{wei2020combating}. Specifically, for CIFAR-10 and CIFAR-100, we use a 7-layer CNN network architecture. For Clothing1M, we use a 18-layer ResNet.

We use the Adam optimizer with a momentum of 0.9 for all experiments. For CIFAR-10 and CIFAR-100, the network is trained for 200 epochs. We set the initial learning rate as 0.001 and the batch size as 128. For Clothing1M, the network is trained for 15 epochs. We set the constant learning rate as $5\times {{10}^{-4}}$ and the batch size as 64. In all experiments, we take the horizontal flipping as the transformation operation, and the ensemble momentum $\gamma $ in Eq.~\ref{ensembling} is set to 0.999. The hyperparameters of total loss weight $\lambda$ and linear drop rate ${{T}_{k}}$ for all experiments are given in Table~\ref{hyperparameters}.

\begin{table}[]
\centering
\caption{\centering The hyperparameters of total loss weight $\lambda$ and linear drop rate ${{T}_{k}}$ for all experiments.}
\begin{tabular}{c|c|c|c}
\hline
Dataset                                     & Flipping-Rate   & $\lambda$ & ${{T}_{k}}$ \\ \hline
\multicolumn{1}{c|}{\multirow{4}{*}{CIFAR-10}} & Symmetry-20\% & {\quad0.8\quad}  & {\quad10\quad}   \\ \cline{2-4} 
\multicolumn{1}{c|}{}                         & Symmetry-50\% & {0.5}  & {20}  \\ \cline{2-4} 
\multicolumn{1}{c|}{}                         & Symmetry-80\% & {0.3}  & {40}   \\ \cline{2-4} 
\multicolumn{1}{c|}{}                         & Asymmetry-40\% & {0.7}  & {50}  \\ \hline
\multirow{4}{*}{CIFAR-100}                     & Symmetry-20\% & {0.8}  & {40}  \\ \cline{2-4} 
                                              & Symmetry-50\% &{0.8}   & {30}  \\ \cline{2-4} 
                                              & Symmetry-80\% & {0.2}  & {30}  \\ \cline{2-4} 
                                              & Asymmetry-40\% & {0.9}  &{30}   \\ \hline
\multicolumn{2}{c|}{Clothing1M}                   & {0.6}  & {5}   \\ \hline
\end{tabular}
\label{hyperparameters}
\end{table}

Following previous works~\cite{han2018co}~\cite{yu2019does}~\cite{wei2020combating}, we verify our method with two types of label noise: \emph{symmetric} and \emph{asymmetric}. The symmetric label noise is generated by using a random one-hot vector to replace the ground-truth label of a sample. The asymmetric label noise is designed to mimic the structure of real-world label noise, such as CAT$\leftrightarrow$DOG, BIRD$\leftrightarrow$AIRPLANE. In each experiment, we also follow the same setting to assume the noise rate $r$ is known, therefore the $\tau$ in Eq. (\ref{update rt}) can be obtained. Specifically, for symmetric label noise, $\tau$ equals to $r$ (\ie, $\tau = r$). For asymmetric label noise, due to the fact that the actual noise rate  $\tau$ in the whole dataset is half of the noisy rate in the noisy classes, we set $\tau = r/2$.

\textbf{Measurements.} We use the test accuracy and label prediction to measure the performance, \ie, \emph{\textbf{test accuracy} = (\# of correct predictions) / (\# of test dataset)} and \emph{\textbf{label precision} = (\# of clean labels) / (\# of all selected labels)}. Intuitively, the algorithm with higher label precision is more robust to the label noise. All experiments are repeated five times and the error bar for STD in each figure has been highlighted as a shade.

\textbf{Baseline.} We use the conventional training with the cross-entropy loss on noisy datasets (abbreviated as Standard) as our baseline and compare our method with following state-of-the-art approaches:

\begin{enumerate}
    \item [-] \textbf{Decoupling}~\cite{malach2017decoupling}, which trains two networks simultaneously, and then updates models only using the instances that have different predictions from these two networks.
    \item [-] \textbf{Co-teaching}~\cite{han2018co}, which trains two networks simultaneously, and each network uses its small-loss instances to update its peer network's parameters.
    \item [-] \textbf{Co-teaching+}~\cite{yu2019does}, which trains two networks simultaneously using the disagreement-update step (data update) and cross-update step (parameters update).
    \item [-] \textbf{JoCoR}~\cite{wei2020combating}, which trains two networks with a pseudo-siamese paradigm and update their parameters simultaneously by a joint loss.
\end{enumerate}

\begin{figure*}[htbp]
\centering
\subfigure{
\begin{minipage}[t]{1\linewidth}
\centering
\includegraphics[width=6in]{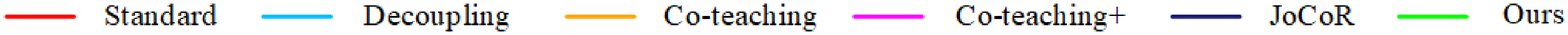}
\end{minipage}
}

\subfigure{
\begin{minipage}[t]{0.23\linewidth}
\centering
\includegraphics[width=1.5in]{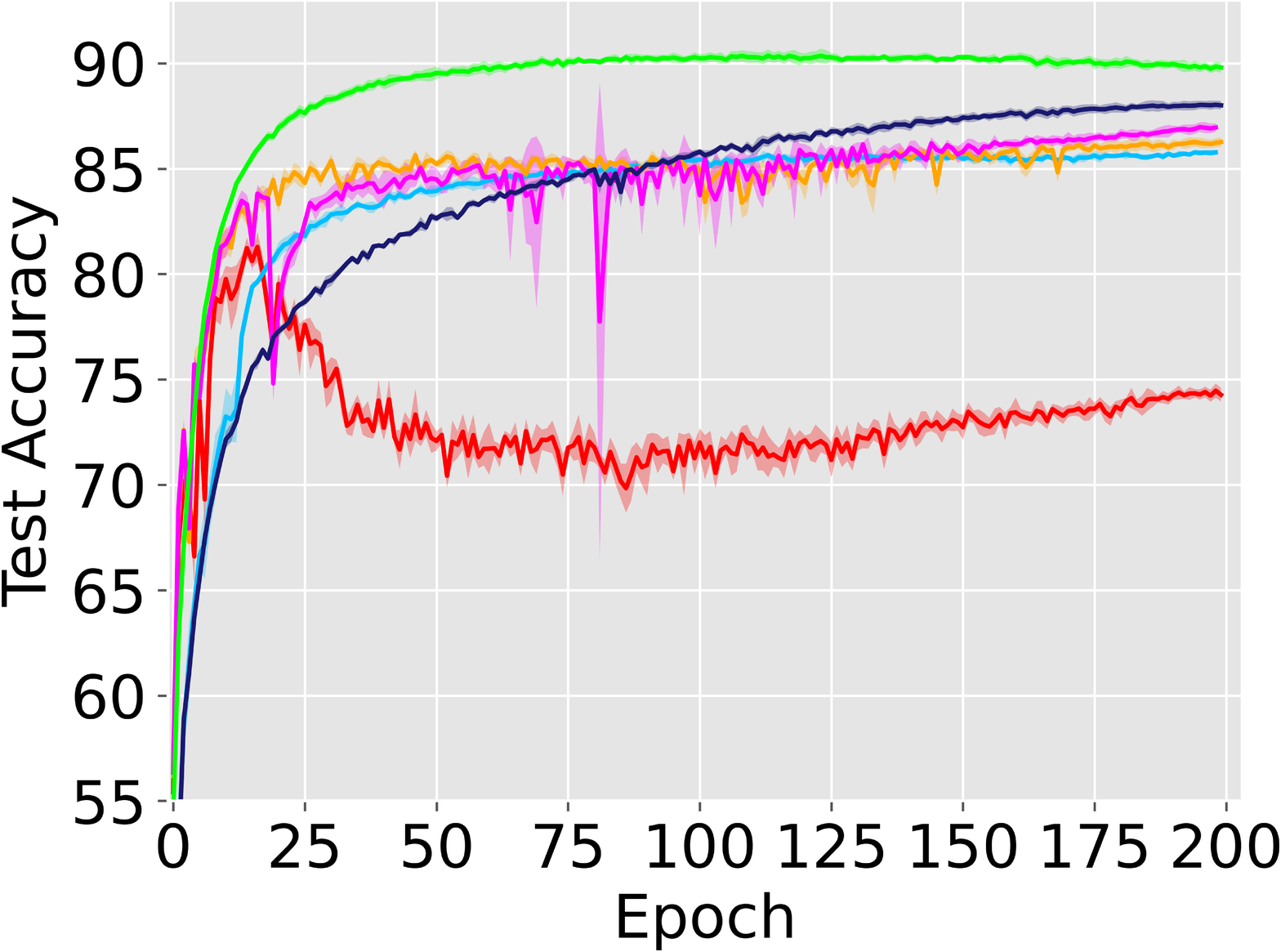}
\end{minipage}
}
\subfigure{
\begin{minipage}[t]{0.23\linewidth}
\centering
\includegraphics[width=1.5in]{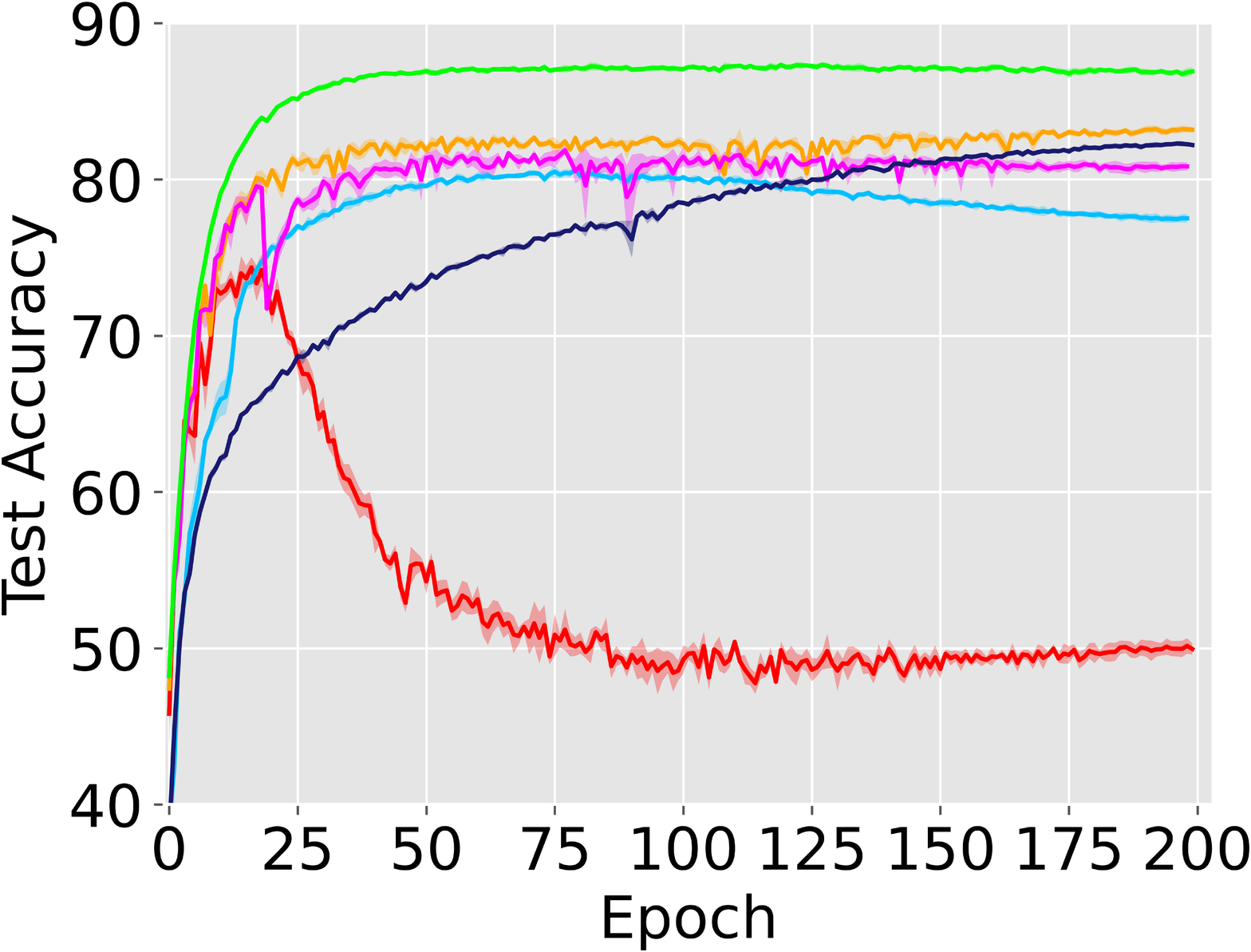}
\end{minipage}
}
\subfigure{
\begin{minipage}[t]{0.23\linewidth}
\centering
\includegraphics[width=1.5in]{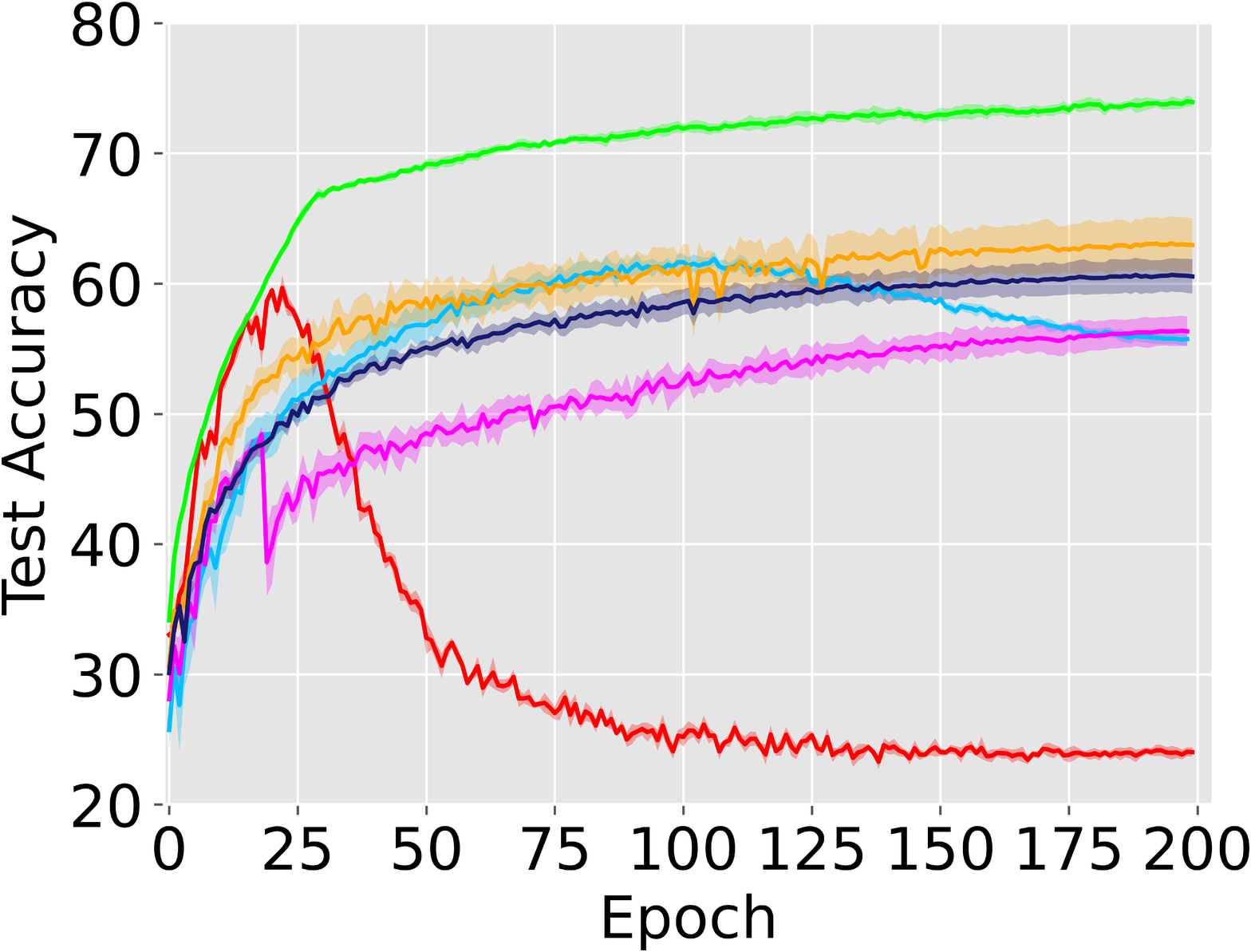}
\end{minipage}
}
\subfigure{
\begin{minipage}[t]{0.23\linewidth}
\centering
\includegraphics[width=1.5in]{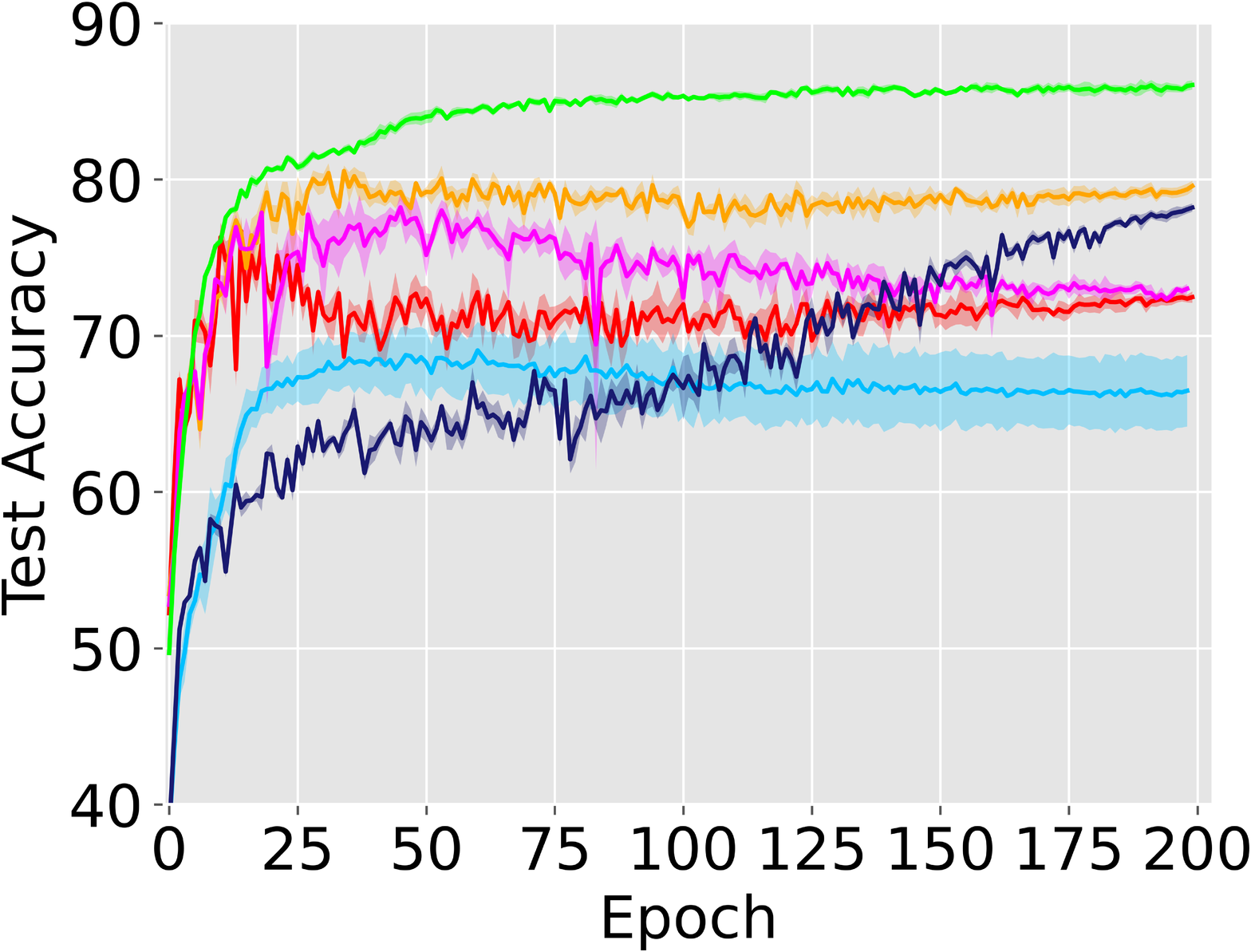}
\end{minipage}
}

\subfigure[(a) Symmetry-20\%]{
\begin{minipage}[t]{0.23\linewidth}
\centering
\includegraphics[width=1.5in]{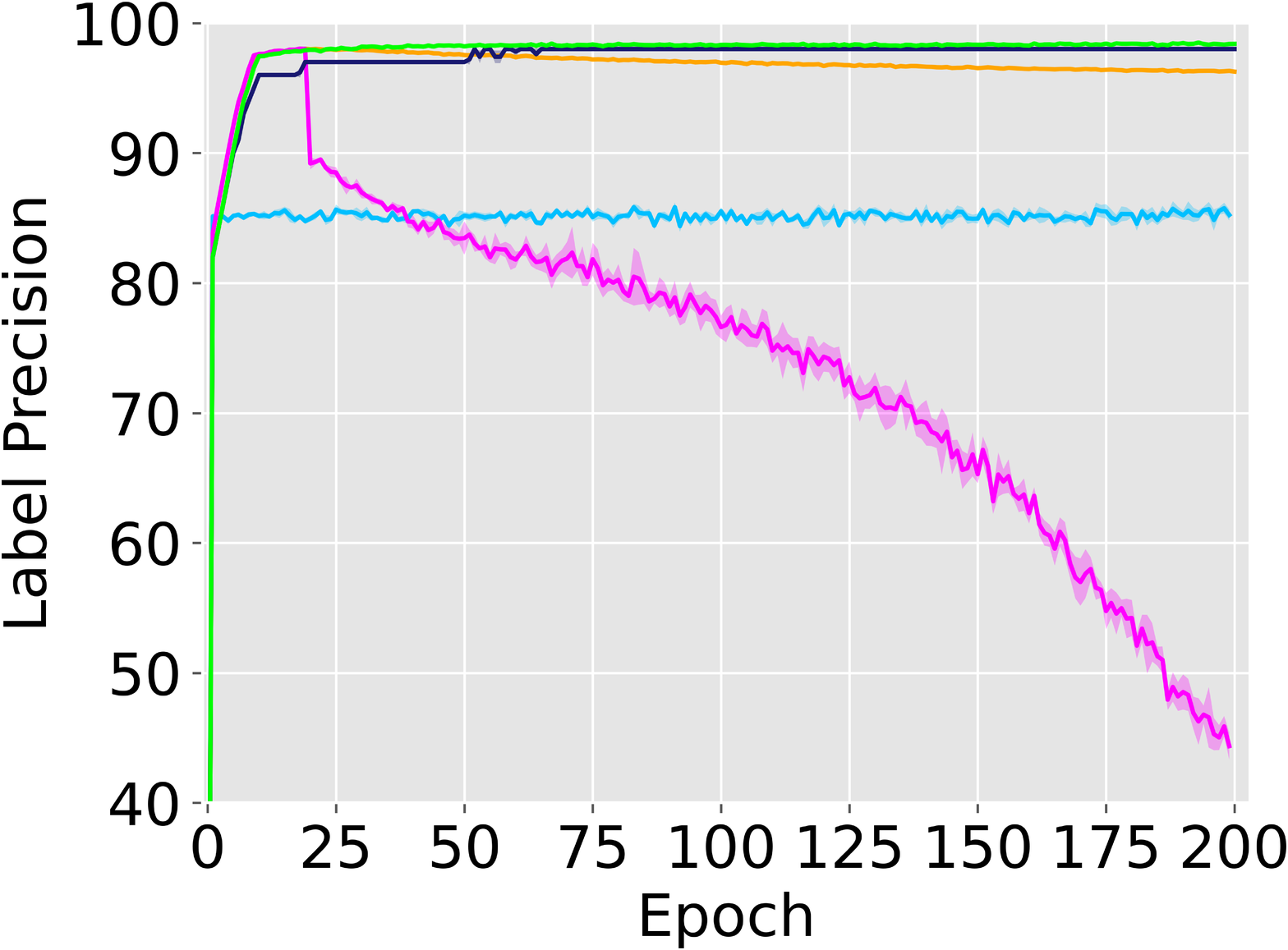}
\end{minipage}
}
\subfigure[(b) Symmetry-50\%]{
\begin{minipage}[t]{0.23\linewidth}
\centering
\includegraphics[width=1.5in]{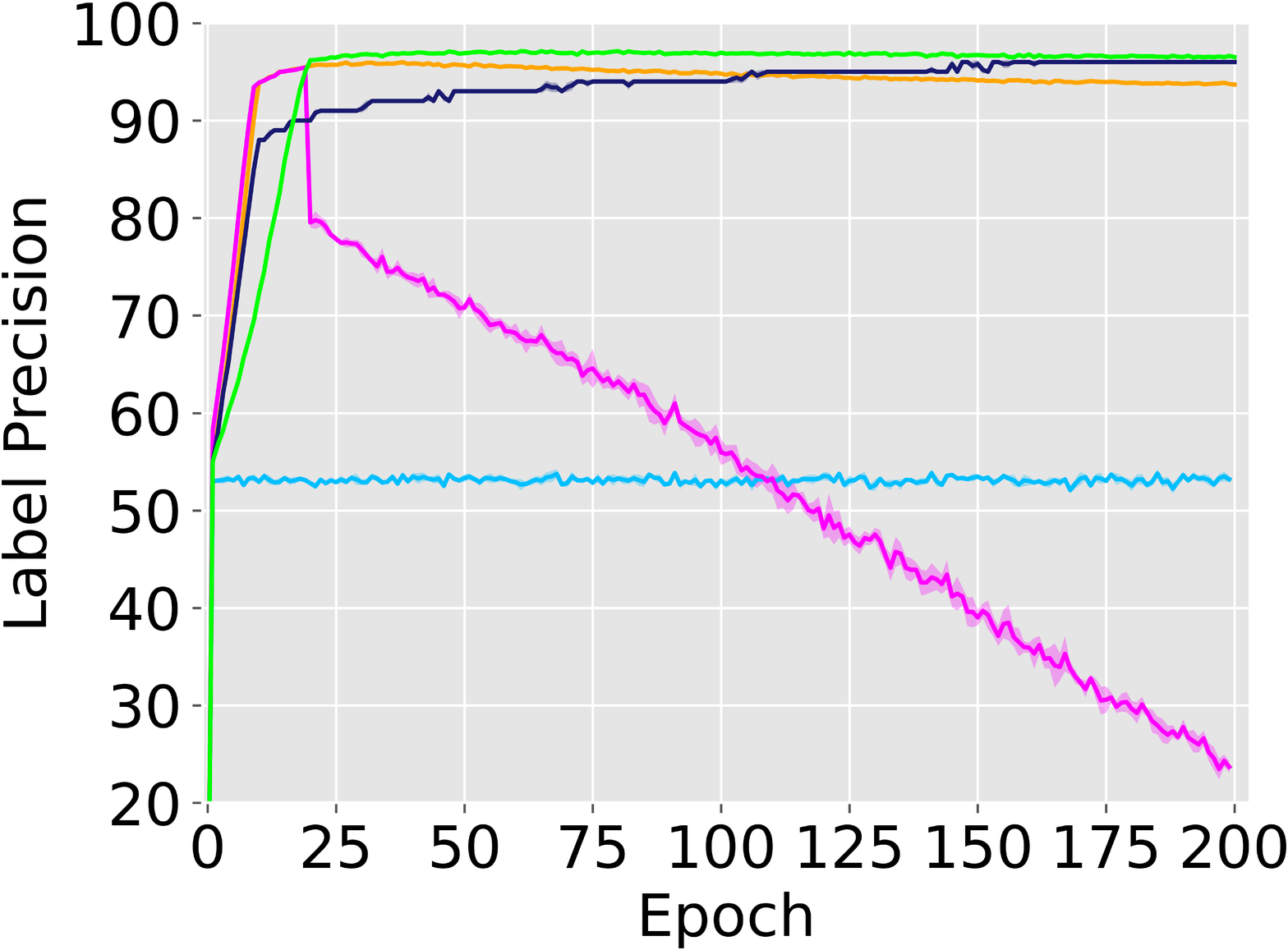}
\end{minipage}
}
\subfigure[(c) Symmetry-80\%]{
\begin{minipage}[t]{0.23\linewidth}
\centering
\includegraphics[width=1.5in]{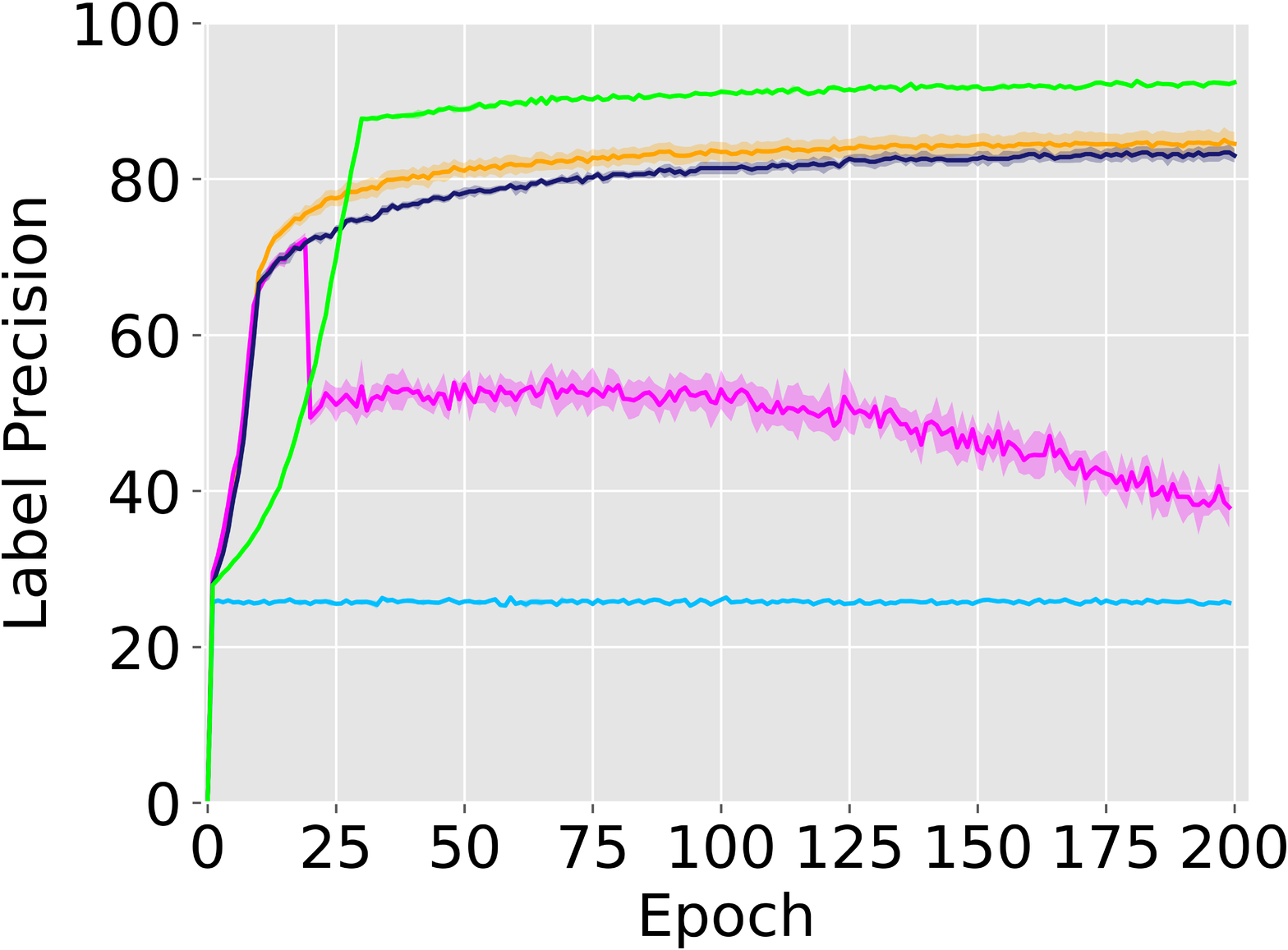}
\end{minipage}
}
\subfigure[(d) Asymmetry-40\%]{
\begin{minipage}[t]{0.23\linewidth}
\centering
\includegraphics[width=1.5in]{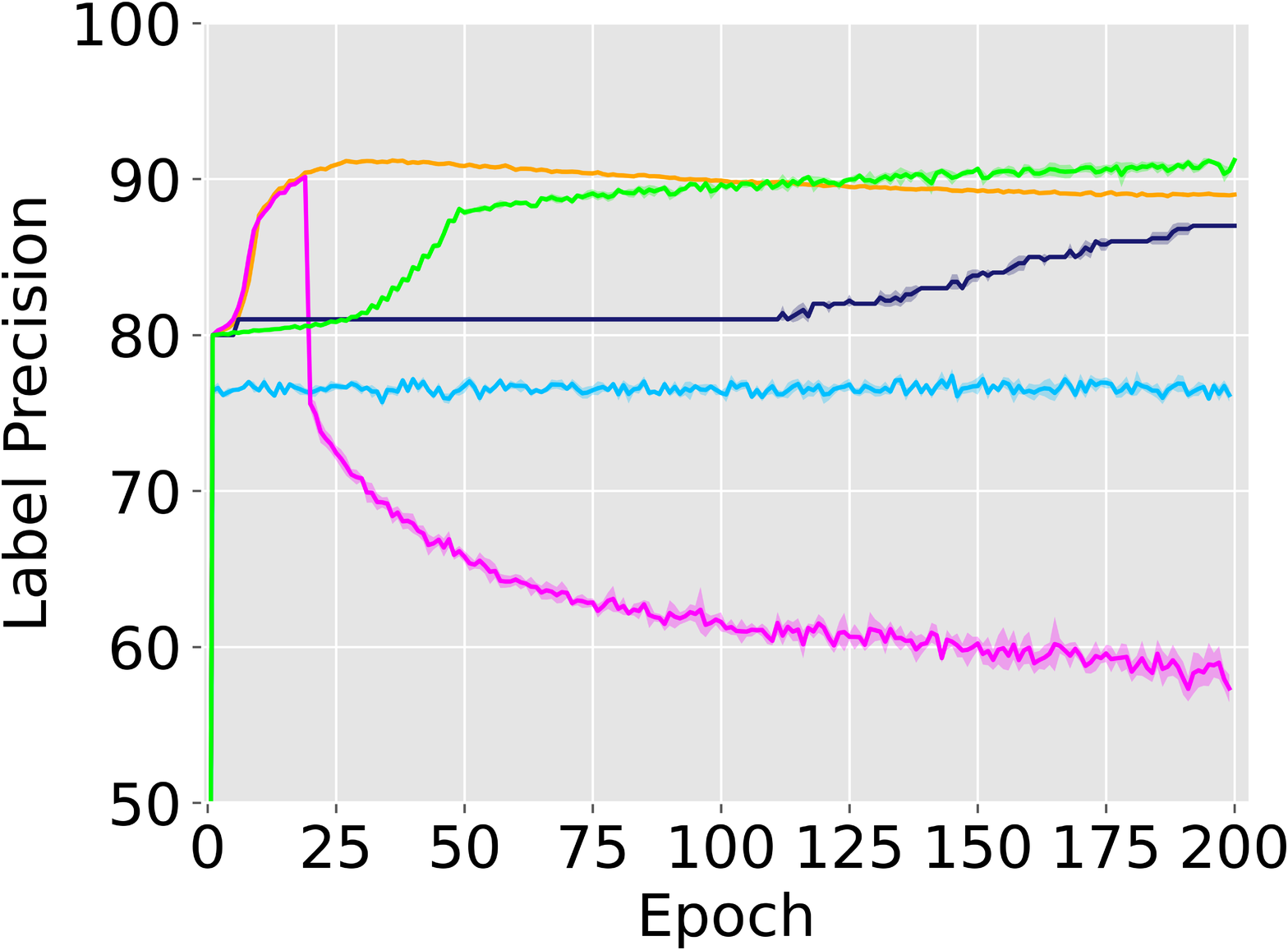}
\end{minipage}
}
\caption{Results on CIFAR-10 dataset. Top: test accuracy (\%) vs. epochs; bottom: label precision (\%) vs. epochs.}
\label{CIFAR10_Accuracy_Figure}
\end{figure*}

\begin{table*}[t]
\renewcommand\arraystretch{1.2}
\centering
\caption{\centering Average test accuracy (\%) on CIFAR-10 over the last 10 epochs.}
\begin{tabular}{l|c|c|c|c|c|c}
\hline\hline
Flipping-Rate  & Standard & Decoupling & Co-teaching & Co-teaching+ & JoCoR & Ours \\ \hline
Symmetry-20\%  & $74.32\pm 0.07$   & $85.72\pm 0.05$    & $86.21\pm 0.04$  & $86.88\pm 0.06$  & $88.00\pm 0.02$  & $\textbf{89.85}\pm \textbf{0.06}$     \\ \hline
Symmetry-50\%  & $49.98\pm 0.08$   & $77.50\pm 0.03$    & $83.15\pm 0.06$  & $80.80\pm 0.03$  & $82.23\pm 0.04$  & $\textbf{86.85}\pm \textbf{0.05}$      \\ \hline
Symmetry-80\%  & $24.03\pm 0.09$   & $55.80\pm 0.07$    & $63.01\pm 0.04$  & $56.31\pm 0.04$  & $60.59\pm 0.05$  & $\textbf{73.84}\pm \textbf{0.08}$      \\ \hline
Asymmetry-40\% & $72.29\pm 0.15$   & $66.34\pm 0.12$    & $79.23\pm 0.17$  & $72.74\pm 0.26$  & $77.81\pm 0.21$  & $\textbf{85.87}\pm \textbf{0.11}$      \\ \hline\hline
\end{tabular}
\label{CIFAR10_Accuracy_Table}
\end{table*}

\subsection{Comparison with state-of-the-art methods}
\subsubsection{Results on CIFAR-10 dataset}

The test accuracy on CIFAR-10 dataset is shown in Table~\ref{CIFAR10_Accuracy_Table}. From the comparison results, we can clearly see that our method performs the best in all four cases. Specifically, in the easiest Symmetry-20\% case, all methods work well, but our method still achieves an improvement more than 1.85\% over the best baseline method JoCoR (89.85\% vs. 88.00\%). When the noise rate raises to 50\%, the test accuracy of Decoupling decreases lower than 80\%, but the other three methods remain above 80\%, where Co-teaching performs better than JoCoR and Co-teaching+. Our method exceeds Co-teaching by 3.70\% (86.85\% vs. 83.15\%). In the Symmetry-80\% case, which means that the network trains with extremely noisy labels, methods based on ``disagreement'' strategy, \ie, Decoupling and Co-teaching+,  cannot work well in this case. Co-teaching and JoCoR only reach to 63.01\% and 60.59\% respectively. Surprisingly, our method largely exceeds these two methods by 10.83\% and 13.25\% respectively. In the hardest Asymmetry-40\%, Decoupling and Co-teaching+ also fail, where Decoupling works much worse than the Standard method. In contrast, Co-teaching, JoCoR and our method perform better. Among them, our method is still the best and exceeds the second best method Co-teaching by 6.64\% (85.87\% vs. 79.23\%).

In the top of Figure~\ref{CIFAR10_Accuracy_Figure}, we show test accuracy vs. number of epochs, which can help us to clearly see the memorization effects of DNNs. For example, the Standard method first learns from clean samples in initial stages and then gradually adapts to noisy ones, therefore the curve of test accuracy first reaches a high level and then decreases gradually. While for Decoupling, the curve of test accuracy first rises from 0 to 100 epochs, then gradually decreases from 100 epochs in two cases (\ie, Symmetry-50\% and Symmetry-80\%). In other two cases (\ie, Symmetry-20\% and Asymmetry-40\%), the test accuracy gradually increases and finally stabilizes. The test accuracy curves of the other four methods follow the similar trends, \ie, they first increase and then level off, but the Co-teaching+ drops slightly in Asymmetry-40\%.

In order to further explain such good performance, we plot the label precision vs. number of epochs on the bottom of Figure~\ref{CIFAR10_Accuracy_Figure}. Only Decoupling, Co-teaching, Co-teaching+, JoCoR and our method are considered here because these algorithms contain the operation of clean instances selection. We can see that the label precision curves of Decoupling and Co-teaching+ fail to select the clean instances in all four cases. The label precision curve of Co-teaching+ first increases at a high level and then gradually decreases, while for Decoupling, the label precision is always at a lower level. These curves show that the ``disagreement'' strategy cannot deal with noisy labels at all because this strategy does not utilize the memorization effects of DNNs during training. In contrast, the label precision curves of the other three methods increase during the training, and then remain a high precision, which means that they can pick up clean instances successfully. Among them, our method can achieve a higher label precision only after training a few epochs, and has always provided more accurate supervisions for the subsequent training process, resulting in a best classification performance.

\subsubsection{Results on CIFAR-100 dataset}
Table~\ref{CIFAR100_Accuracy_Table} shows the test accuracy on CIFAR-100 dataset. Different from CIFAR-10 dataset, CIFAR-100 dataset is more difficult for noisy-labeled image classification because it contains 100 classes. Similarly, the test accuracy of our method on CIFAR-100 dataset achieves the excellent performance again. Specifically, in the easiest Symmetry-20\% case, all methods work well, but our method exceeds the second best method JoCoR 4.85\% (62.00\% vs. 57.15\%). When the noise rate raises to 50\%, our method is still the best and outperforms the second method JoCoR 6.59\% (55.54\% vs. 48.95\%). In the extreme Symmetry-80\% case, Decoupling, Co-teaching+ and JoCoR fail to achieve good performance, \ie, they are all below 20\%. Co-teaching also only achieves 22.08\%, but our method achieves 34.86\%, which is much higher than Co-teaching by 12.78\%. In the hardest case (\ie, Asymmetry-40\%), Decoupling, Co-teaching and JoCoR are below 35\%. Co-teaching+ achieves the best performance (45.19\%), but our method is comparable with Co-teaching+ (43.56\%). Meanwhile, our method achieves the better performance on label precision than Co-teaching+ in this case. 

\begin{figure*}[htbp]
\centering
\subfigure{
\begin{minipage}[t]{1\linewidth}
\centering
\includegraphics[width=6in]{./tiao_crop.eps}
\end{minipage}
}

\subfigure{
\begin{minipage}[t]{0.23\linewidth}
\centering
\includegraphics[width=1.5in]{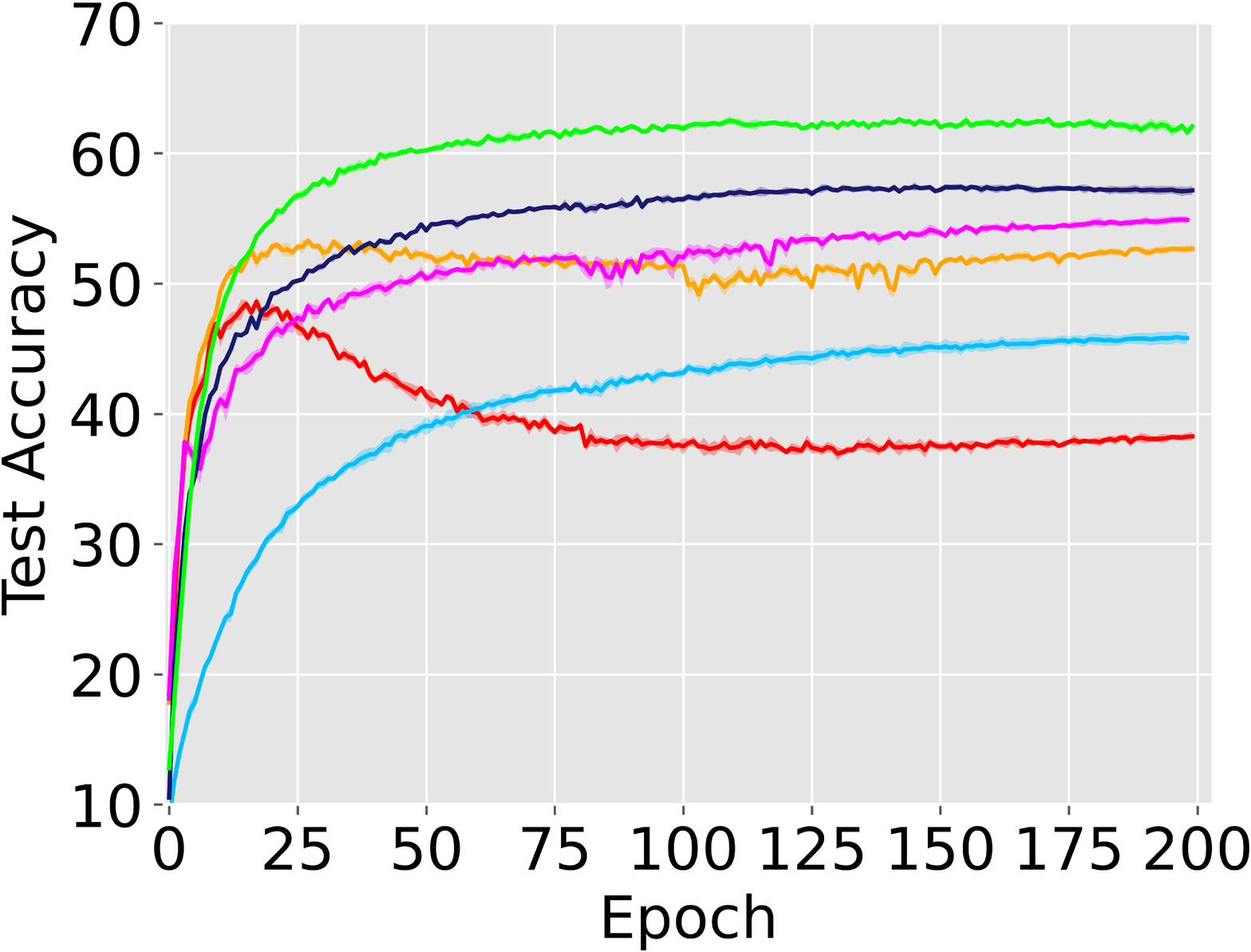}
\end{minipage}
}
\subfigure{
\begin{minipage}[t]{0.23\linewidth}
\centering
\includegraphics[width=1.5in]{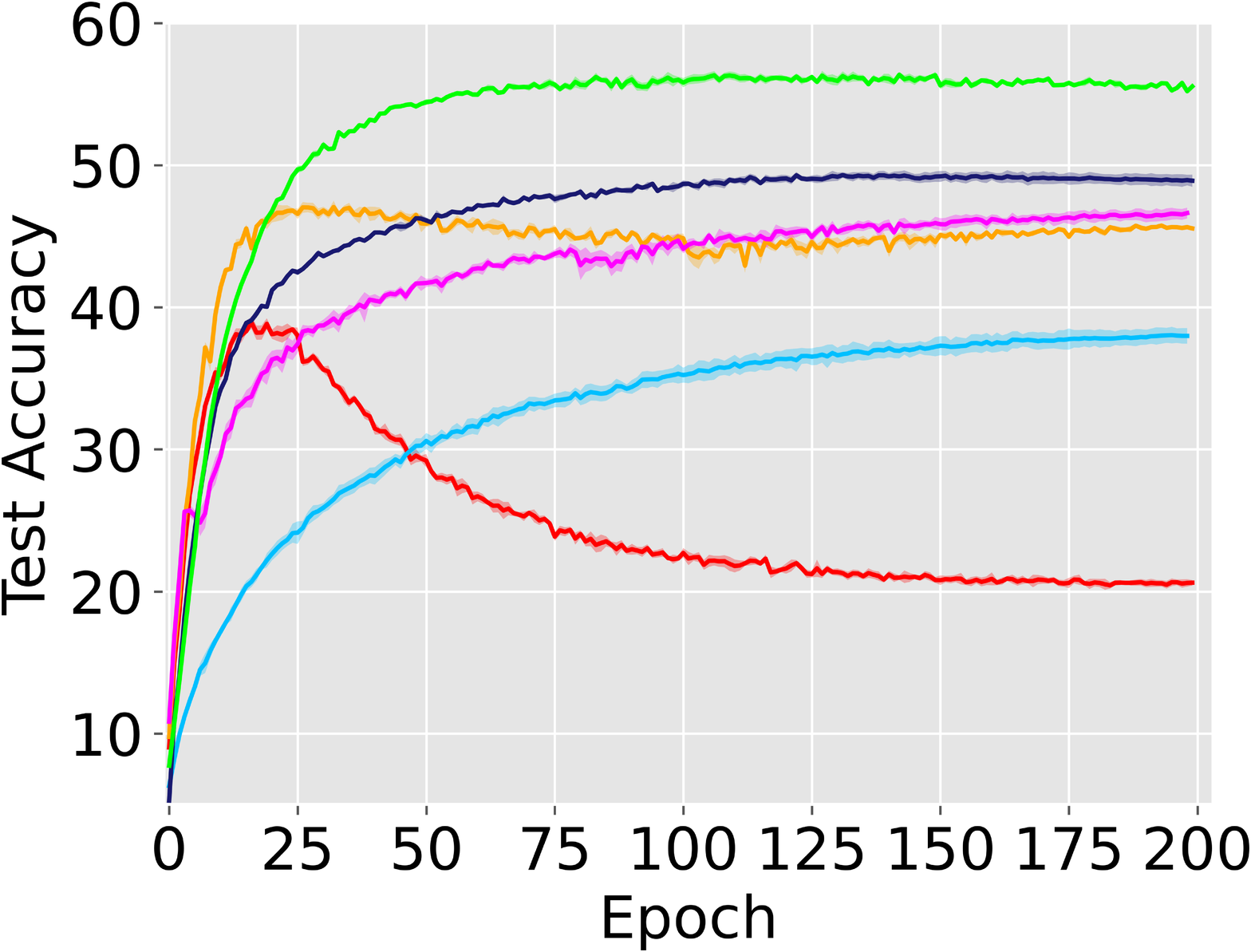}
\end{minipage}
}
\subfigure{
\begin{minipage}[t]{0.23\linewidth}
\centering
\includegraphics[width=1.5in]{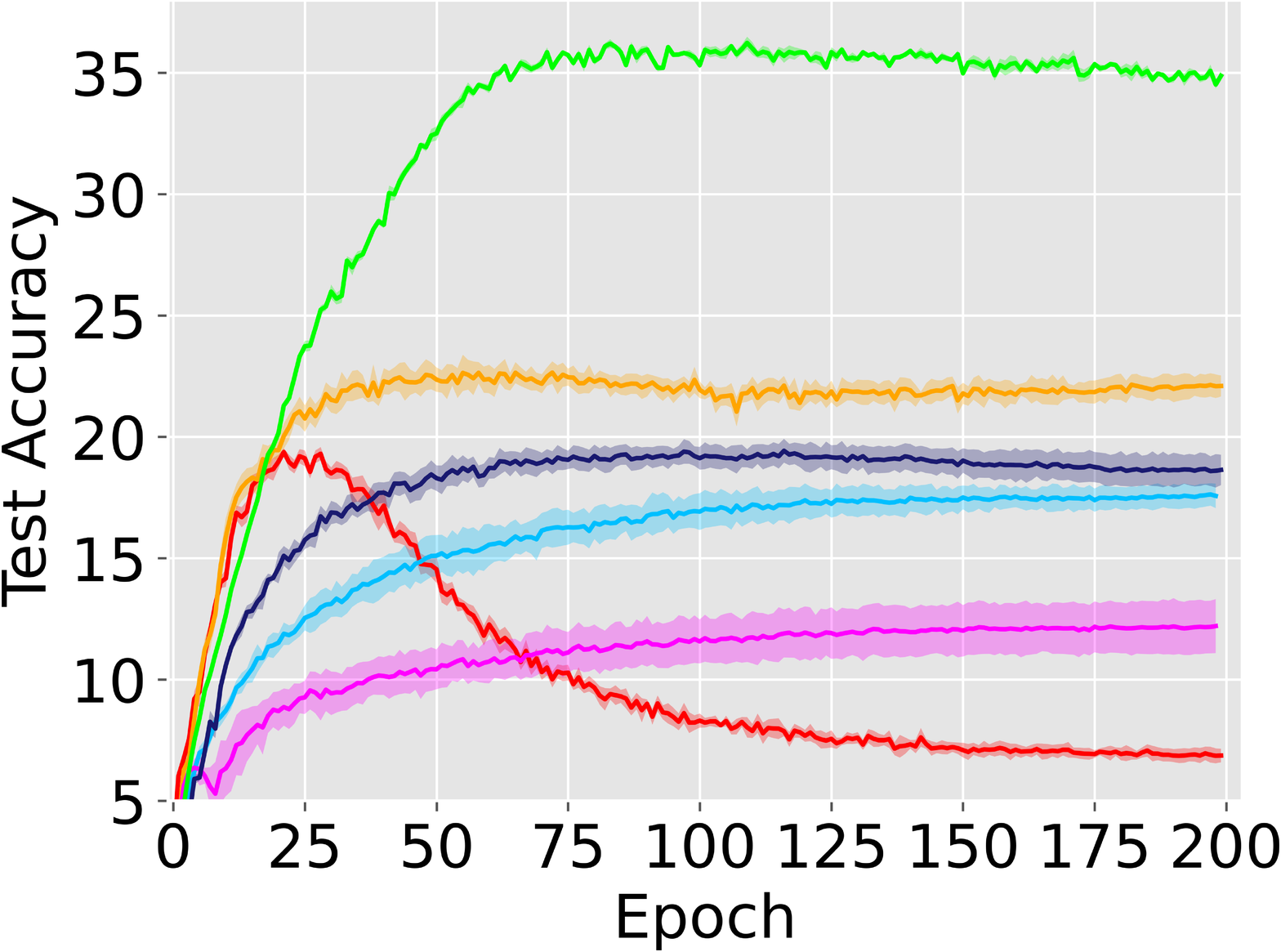}
\end{minipage}
}
\subfigure{
\begin{minipage}[t]{0.23\linewidth}
\centering
\includegraphics[width=1.5in]{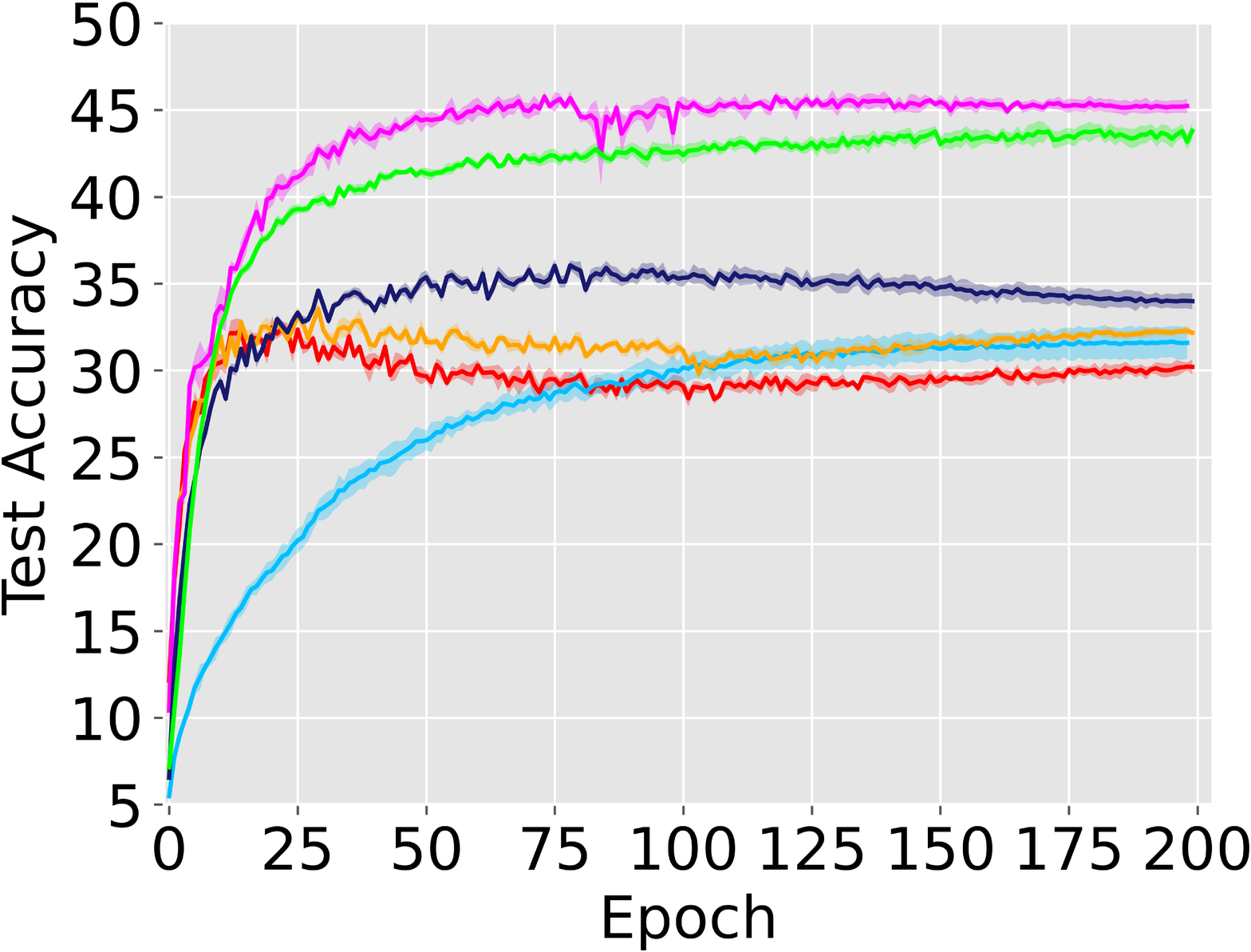}
\end{minipage}
}

\subfigure[(a) Symmetry-20\%]{
\begin{minipage}[t]{0.23\linewidth}
\centering
\includegraphics[width=1.5in]{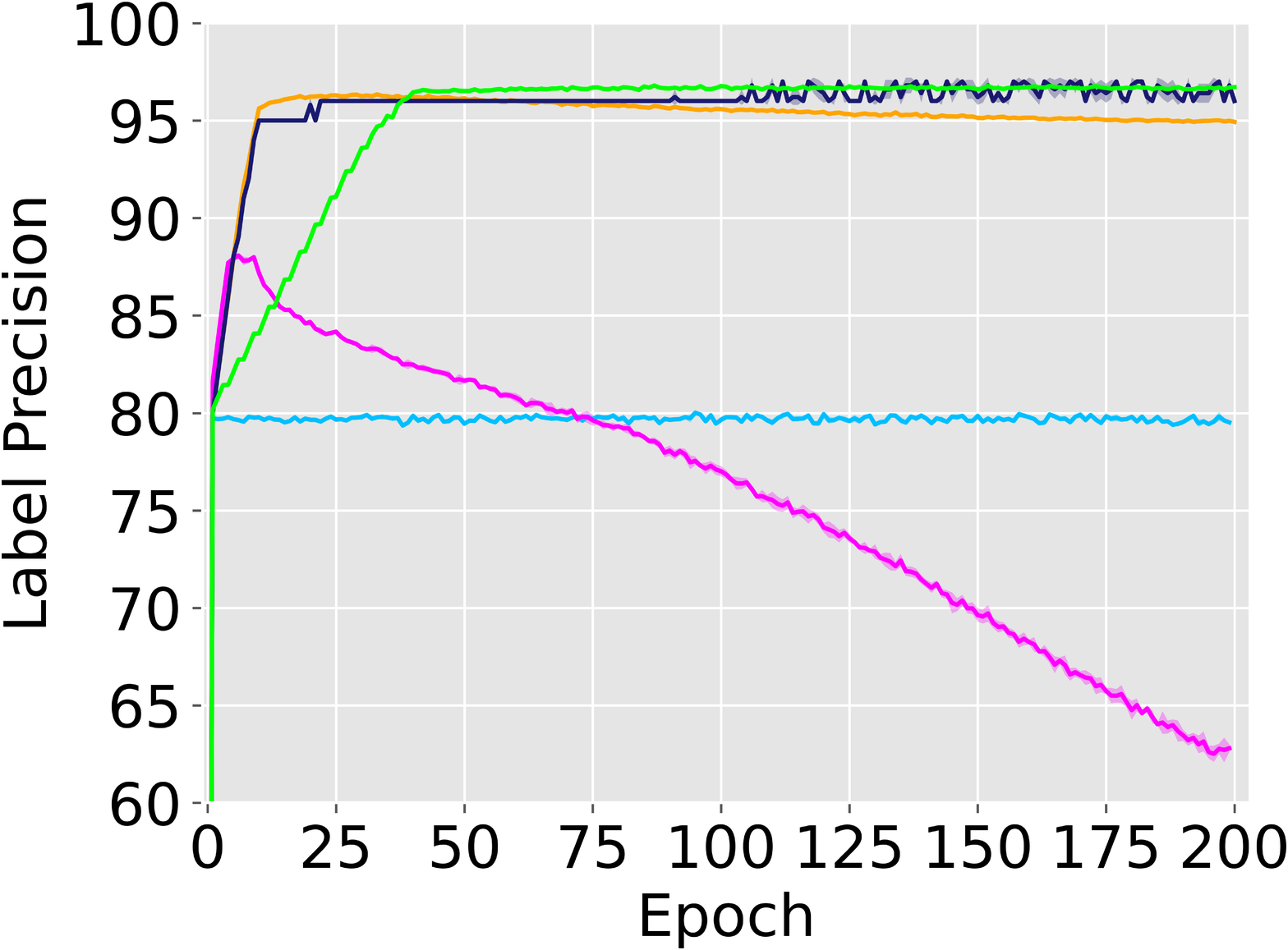}
\end{minipage}
}
\subfigure[(b) Symmetry-50\%]{
\begin{minipage}[t]{0.23\linewidth}
\centering
\includegraphics[width=1.5in]{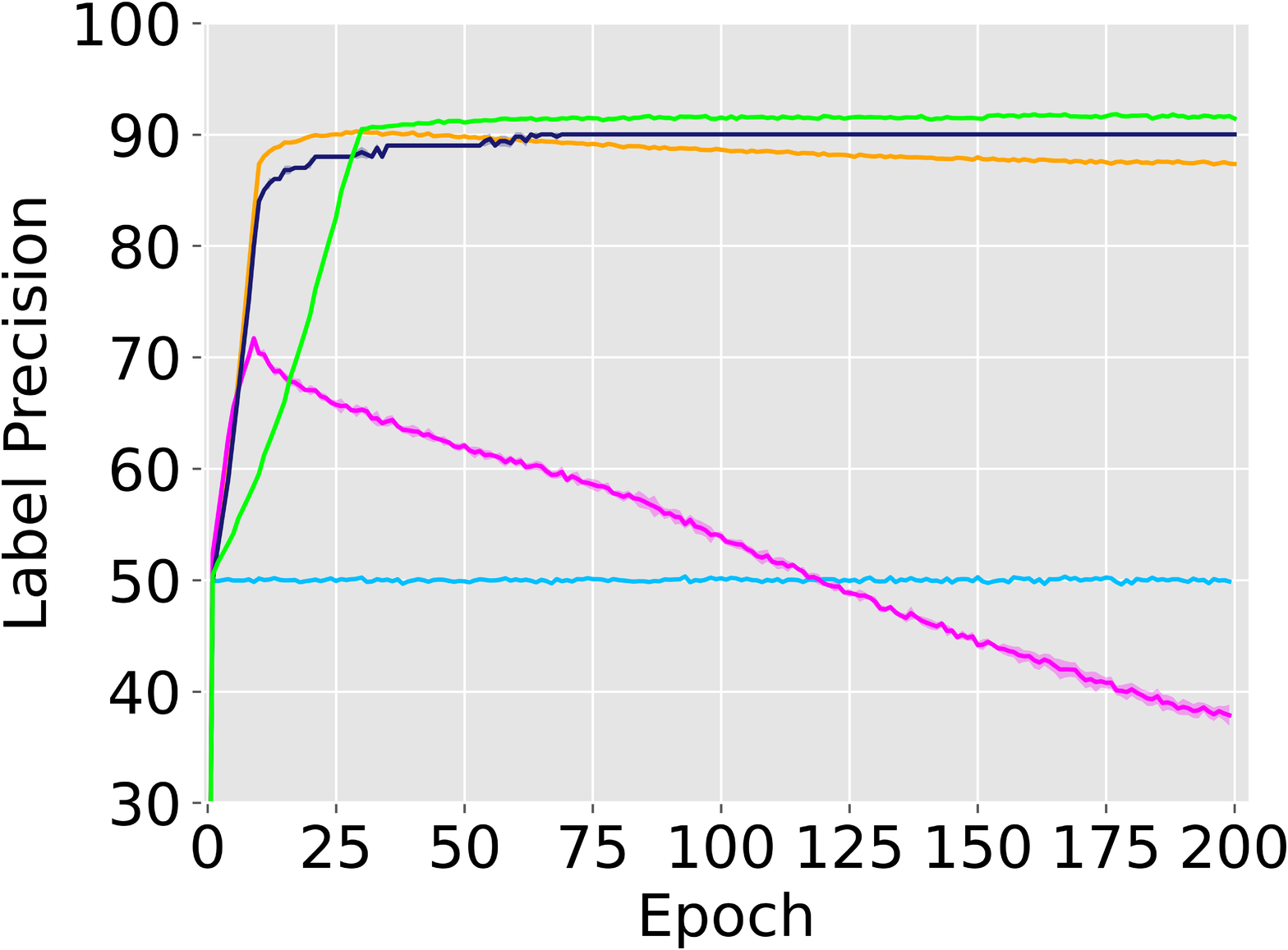}
\end{minipage}
}
\subfigure[(c) Symmetry-80\%]{
\begin{minipage}[t]{0.23\linewidth}
\centering
\includegraphics[width=1.5in]{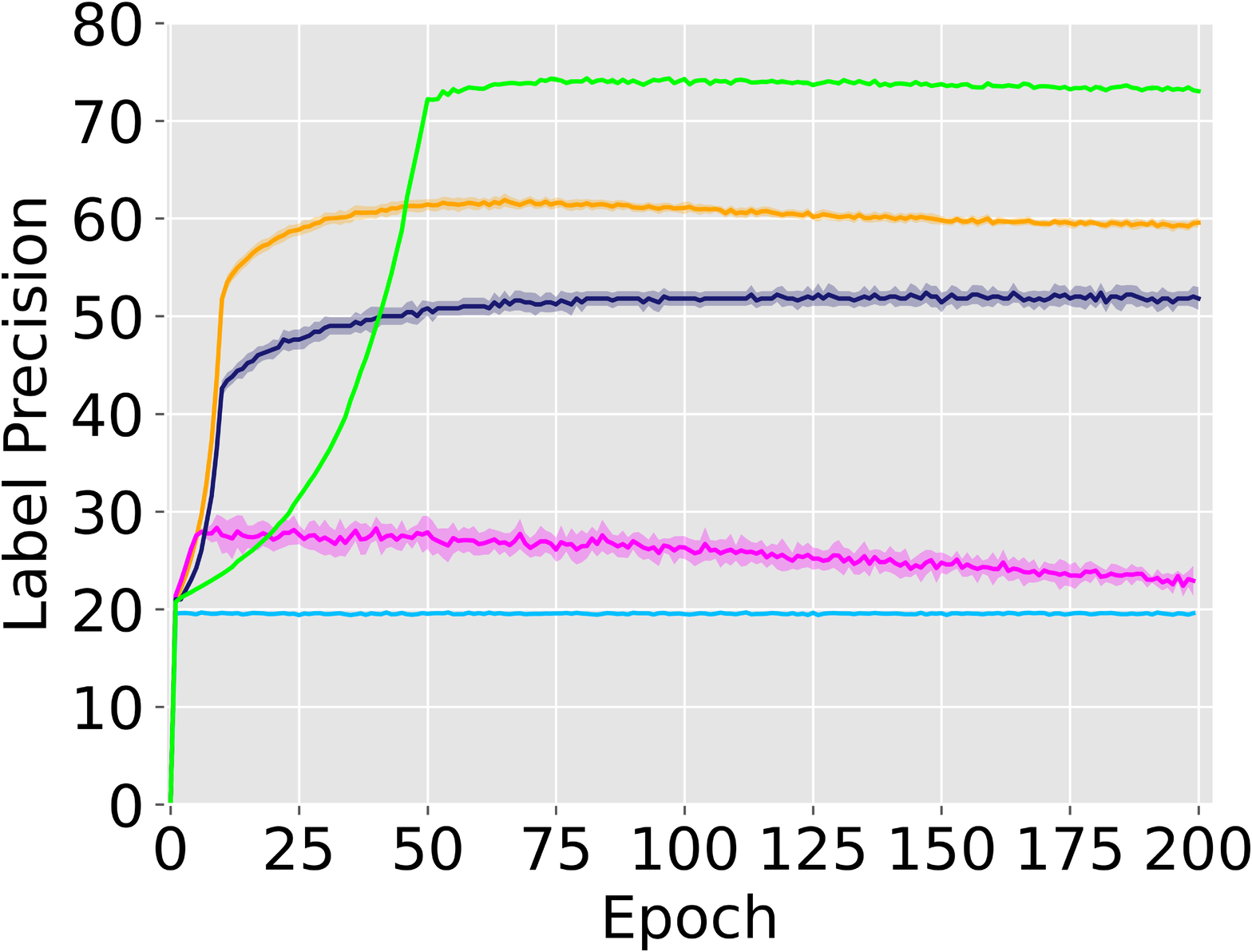}
\end{minipage}
}
\subfigure[(d) Asymmetry-40\%]{
\begin{minipage}[t]{0.23\linewidth}
\centering
\includegraphics[width=1.5in]{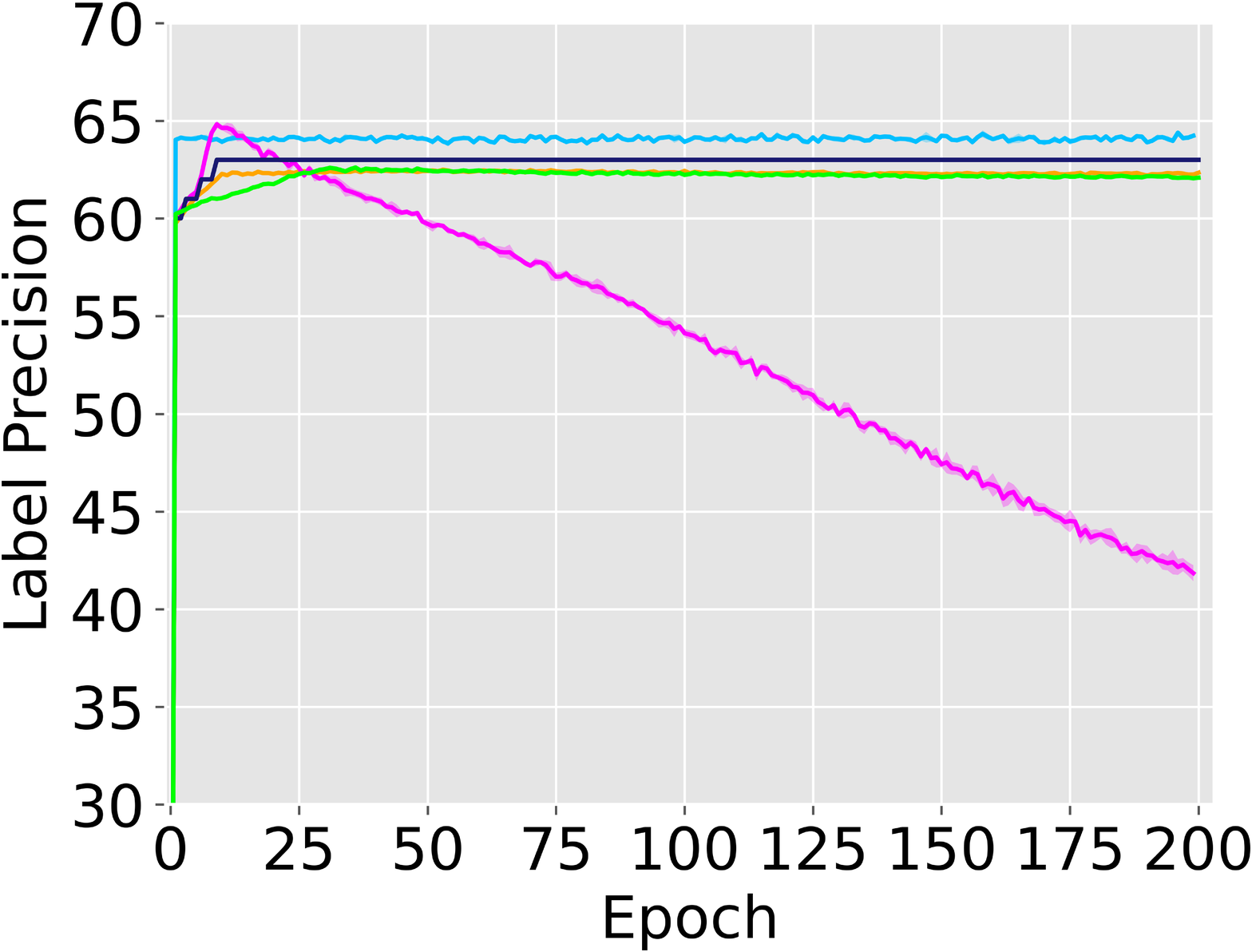}
\end{minipage}
}
\caption{Results on CIFAR-100 dataset. Top: test accuracy (\%) vs. epochs; bottom: label precision (\%) vs. epochs.}
\label{CIFAR100_Accuracy_Figure}
\end{figure*}

\begin{table*}[t]
\renewcommand\arraystretch{1.2}
\centering
\caption{\centering Average test accuracy (\%) on CIFAR-100 over the last 10 epochs.}
\begin{tabular}{l|c|c|c|c|c|c}
\hline\hline
Flipping-Rate  & Standard & Decoupling & Co-teaching & Co-teaching+ & JoCoR & Ours \\ \hline
Symmetry-20\%  & $38.15\pm 0.07$   & $45.80\pm 0.05$    & $52.58\pm 0.10$  & $54.82\pm 0.07$  & $57.15\pm 0.04$  & $\textbf{62.00}\pm \textbf{0.20}$     \\ \hline
Symmetry-50\%  & $20.58\pm 0.05$   & $37.98\pm 0.05$    & $45.62\pm 0.06$  & $46.51\pm 0.07$  & $48.95\pm 0.03$  & $\textbf{55.54}\pm \textbf{0.20}$      \\ \hline
Symmetry-80\%  & $6.88\pm 0.03$    & $17.56\pm 0.04$    & $22.08\pm 0.04$  & $12.16\pm 0.03$  & $18.63\pm 0.03$  & $\textbf{34.86}\pm \textbf{0.17}$      \\ \hline
Asymmetry-40\% & $30.08\pm 0.11$   & $31.60\pm 0.03$    & $32.22\pm 0.04$  & $\textbf{45.19}\pm \textbf{0.04}$  & $34.00\pm 0.04$  & $43.56\pm 0.20$      \\ \hline\hline
\end{tabular}
\label{CIFAR100_Accuracy_Table}
\end{table*}

Figure~\ref{CIFAR100_Accuracy_Figure} shows the test accuracy and label precision vs. epochs. We can observe the memorization effects of DNNs using different methods again. Similar with the results on CIFAR-10 dataset, the Standard method first increases a high level and then decreases gradually. Decoupling and Co-teaching+ always worse than the other three approaches in Symmetry-20\%, 50\% and 80\%, and the label precision curves of them also explain the results that they have no ability to deal with the noisy labels. Instead, Co-teaching, JoCoR and our method mitigate the memorization issue. They can select clean instances out of the noisy ones successfully, resulting in a good performance. Only in Asymmetry-40\%, our method is slightly lower than Co-teaching+, but our method achieves the better performance on label precision than Co-teaching+. Besides, 
by observing the convergence process of the model trained with five methods, we can see that our method can stabilize and achieve the better classification performance only after about 100 epochs.

\subsubsection{Results on Clothing1M dataset}
We demonstrate the efficiency of our proposed method on real-world noisy labels using Clothing1M dataset. The results are shown in Table~\ref{Clothing1M_Accuracy}. We report the \emph{best} test accuracy across all epochs and the \emph{last} test accuracy at the end of training. Our method gains the best performance compared to the state-of-the-art methods in \emph{best} and \emph{last} test accuracy and achieves a significant improvement in \emph{last} accuracy of 3.14\% over Standard method, and an improvement of 0.24\% over the second best method JoCoR.

\begin{table}[t]
\renewcommand\arraystretch{1.1}
\centering
\caption{\centering Classification accuracy (\%) on the Clothing1M test set.}
\begin{tabular}{c|c|c}
\hline\hline
\quad Methods \quad            & {\quad\quad \emph{Best}\quad\quad}      & \quad \emph{Last} \quad\quad \\ \hline
\quad Standard \quad           & {\quad\quad 67.44 \quad\quad}   & {\quad\quad 66.99 \quad\quad}    \\ \hline
\quad Decoupling \quad         & {\quad\quad 68.48 \quad\quad}   & {\quad\quad 67.32 \quad\quad}    \\ \hline
\quad Co-teaching \quad        & {\quad\quad 69.21 \quad\quad}   & {\quad\quad 68.51 \quad\quad}    \\ \hline
\quad Co-teaching+ \quad       & {\quad\quad 59.32 \quad\quad}   & {\quad\quad 58.79 \quad\quad}    \\ \hline
\quad JoCoR  \quad             & {\quad\quad 70.67 \quad\quad}   & {\quad\quad 69.89 \quad\quad}    \\ \hline
\quad Ours \quad               & {\quad\quad \textbf{70.77} \quad\quad}   & {\quad\quad \textbf{70.13} \quad\quad}    \\ \hline\hline
\end{tabular}
\label{Clothing1M_Accuracy}
\end{table}

\subsection{Ablation study}
In this section, we evaluate two components, \ie, transform consistency and soft classification loss, in our proposed method  by conducting ablation studies on CIFAR-10 and CIFAR-100 datasets. Results are shown in Table~\ref{Ablation study}. We create a baseline model that utilizes only off-line hard labels for training (the first row).

\begin{table*}[h]
\renewcommand\arraystretch{1.2}
\centering
\caption{\centering Ablation study results in terms of test accuracy (\%) on CIFAR-10 and CIFAR-100 datasets.}
\begin{tabular}{c|cccc|cccc}
\hline\hline
Dataset & \multicolumn{4}{c|}{CIFAR-10}      & \multicolumn{4}{c}{CIFAR-100}     \\ \hline
Noise type       & \multicolumn{3}{c}{Symmetry} & Asymmetry   & \multicolumn{3}{c}{Symmetry} & Asymmetry   \\ \hline
Methods/Noise ratio       & 20\%      & 50\%     & 80\%    & 40\%     & 20\%      & 50\%    & 80\%    & 40\%     \\ \hline

$\lambda \text{=}0$ \& w/o ${\tilde{\theta }}$       &\tabincell{c}{87.38\\$\pm 0.16$}         &\tabincell{c}{84.21\\$\pm 0.09$}        & \tabincell{c}{72.64\\$\pm 0.12$}      &\tabincell{c}{79.26\\$\pm 0.21$}        &\tabincell{c}{55.52\\$\pm 0.09$}         &\tabincell{c}{48.32\\$\pm 0.17$}       & \tabincell{c}{25.20\\$\pm 0.14$}      &\tabincell{c}{34.79\\$\pm 0.16$}        \\\hline

$\lambda \text{=}0$       &\tabincell{c}{87.92\\$\pm 0.08$}         &\tabincell{c}{84.87\\$\pm 0.05$}        & \tabincell{c}{73.45\\$\pm 0.11$}      &\tabincell{c}{80.25\\$\pm 0.25$}        &\tabincell{c}{55.58\\$\pm 0.17$}         &\tabincell{c}{48.89\\$\pm 0.20$}       &\tabincell{c}{25.99\\$\pm 0.17$}       &\tabincell{c}{35.62\\$\pm 0.22$}        \\\hline

w/o ${\tilde{\theta }}$       & \tabincell{c}{89.23\\ $\pm 0.07$}         &\tabincell{c}{86.21\\$\pm 0.11$}        &\tabincell{c}{72.85\\ $\pm 0.20$}       &\tabincell{c}{84.37\\$\pm 0.38$}        &\tabincell{c}{61.62\\$\pm 0.16$}         &\tabincell{c}{54.09\\$\pm 0.19$}       &\tabincell{c}{33.15\\$\pm 0.13$}       &\tabincell{c}{42.42\\$\pm 0.18$}        \\\hline

Ours       &\tabincell{c}{\textbf{89.85}\\$\pm \textbf{0.06}$}         &\tabincell{c}{\textbf{86.85}\\$\pm \textbf{0.05}$}        &\tabincell{c}{\textbf{73.84}\\$\pm \textbf{0.08}$}       &\tabincell{c}{\textbf{85.87}\\$\pm \textbf{0.11}$}        &\tabincell{c}{\textbf{62.00}\\$\pm \textbf{0.20}$}         & \tabincell{c}{\textbf{55.54}\\$\pm \textbf{0.20}$}      &\tabincell{c}{\textbf{34.86}\\$\pm \textbf{0.17}$}       & \tabincell{c}{\textbf{43.56}\\$\pm \textbf{0.20}$}       \\ \hline\hline
\end{tabular}
\label{Ablation study}
\end{table*}

\begin{table*}[t]
\renewcommand\arraystretch{1.1}
\centering
\caption{\centering Test accuracy (\%) base00d0 on various transforms on CIFAR-10 and CIFAR-100 datasets.}
\begin{tabular}{c|cccc|cccc}
\hline\hline
Dataset & \multicolumn{4}{c|}{CIFAR10}      & \multicolumn{4}{c}{CIFAR100}     \\ \hline
Noise type       & \multicolumn{3}{c}{Symmetry} & Asymmetry   & \multicolumn{3}{c}{Symmetry} & Asymmetry   \\ \hline
Transform/Noise ratio       & 20\%      & 50\%     & 80\%    & 40\%     & 20\%      & 50\%    & 80\%    & 40\%     \\ \hline

Scaling       &\tabincell{c}{83.06\\$\pm 0.19$}         &\tabincell{c}{81.78\\$\pm 0.17$}        &\tabincell{c}{69.63\\$\pm 0.28$}       &\tabincell{c}{71.96\\$\pm 0.41$}        &\tabincell{c}{49.52\\$\pm 0.37$}         & \tabincell{c}{43.53\\$\pm 0.29$}      &\tabincell{c}{22.72\\$\pm 0.25$}       & \tabincell{c}{34.18\\$\pm 0.33$}       \\ \hline

Rotation-$90{}^\circ $       & \tabincell{c}{86.51\\$\pm 0.14$}         &\tabincell{c}{83.02\\$\pm 0.14$}        &\tabincell{c}{68.70\\$\pm 0.15$}      &\tabincell{c}{78.77\\$\pm 0.26$}  &\tabincell{c}{54.97\\$\pm 0.28$}        &\tabincell{c}{49.40\\$\pm 0.25$}         &\tabincell{c}{29.13\\$\pm 0.31$}              &\tabincell{c}{35.93\\$\pm 0.28$}        \\\hline

Rotation-$180{}^\circ $       &\tabincell{c}{87.23\\$\pm 0.08$}         &\tabincell{c}{82.36\\$\pm 0.22$}        & \tabincell{c}{67.24\\$\pm 0.28$}      &\tabincell{c}{78.51\\$\pm 0.16$}        &\tabincell{c}{57.46\\$\pm 0.25$}         &\tabincell{c}{51.77\\$\pm 0.26$}       &\tabincell{c}{32.88\\$\pm 0.17$}       &\tabincell{c}{40.21\\$\pm 0.27$}        \\\hline

Rotation-$270{}^\circ $       &\tabincell{c}{86.38\\$\pm 0.16$}         &\tabincell{c}{83.09\\$\pm 0.15$}        & \tabincell{c}{68.37\\$\pm 0.10$}      &\tabincell{c}{77.86\\$\pm 0.19$}        &\tabincell{c}{54.83\\$\pm 0.25$}         &\tabincell{c}{48.70\\$\pm 0.26$}       & \tabincell{c}{29.56\\$\pm 0.29$}      &\tabincell{c}{35.87\\$\pm 0.35$}        \\\hline

Rotation-$360{}^\circ $       &\tabincell{c}{87.31\\$\pm 0.12$}         &\tabincell{c}{83.33\\$\pm 0.16$}        &\tabincell{c}{71.63\\$\pm 0.30$}       &\tabincell{c}{80.29\\$\pm 0.22$}        &\tabincell{c}{56.99\\$\pm 0.24$}         & \tabincell{c}{50.53\\$\pm 0.30$}      &\tabincell{c}{31.41\\$\pm 0.28$}       & \tabincell{c}{41.87\\$\pm 0.22$}       \\ \hline

Vertical flipping       &\tabincell{c}{86.13\\$\pm 0.11$}         &\tabincell{c}{80.59\\$\pm 0.17$}        &\tabincell{c}{67.03\\$\pm 0.17$}       &\tabincell{c}{78.25\\$\pm 0.16$}        &\tabincell{c}{55.95\\$\pm 0.22$}         & \tabincell{c}{49.09\\$\pm 0.21$}      &\tabincell{c}{30.09\\$\pm 0.29$}       & \tabincell{c}{38.07\\$\pm 0.32$}       \\ \hline

Horizontal flipping       &\tabincell{c}{\textbf{89.85}\\$\pm \textbf{0.06}$}         &\tabincell{c}{\textbf{86.85}\\$\pm \textbf{0.05}$}        &\tabincell{c}{\textbf{73.84}\\$\pm \textbf{0.08}$}       &\tabincell{c}{\textbf{85.87}\\$\pm \textbf{0.11}$}        &\tabincell{c}{\textbf{62.00}\\$\pm \textbf{0.20}$}         & \tabincell{c}{\textbf{55.54}\\$\pm \textbf{0.20}$}      &\tabincell{c}{\textbf{34.86}\\$\pm \textbf{0.17}$}       & \tabincell{c}{\textbf{43.56}\\$\pm \textbf{0.20}$}       \\ \hline\hline
\end{tabular}
\label{transform}
\end{table*}

\textbf{Effectiveness of transform consistency.} To observe the effectiveness of the KL loss, we set $\lambda=0$ in Eq.\ref{total loss}. The results are shown in the second row of Table~\ref{Ablation study} (denoted as $\lambda=0$). The test accuracy of four cases drops by 0.39\% to 5.62\% on CIFAR-10 dataset and 6.42\% to 8.87\% on CIFAR-100 dataset. These experiments fully demonstrate that the clean and noisy samples can be distinguished more easily by adding KL loss and further verify the effectiveness of transform consistency.

\textbf{Effectiveness of soft classification loss.} In order to investigate the effectiveness of the soft classification loss, we perform the experiment by only utilizing the off-line hard labels as the supervisory information (\ie, $\mathcal{L}_{c}^{s}({{x}_{i}})=0$ in Eq.~\ref{soft_loss1} and $\mathcal{L}_{c}^{s}({{\tilde{x}}_{i}})=0$ in Eq.~\ref{soft_loss2}). The performance is presented in the third row of Table~\ref{Ablation study} (denoted as w/o ${\tilde{\theta }}$). The test accuracy of four cases drops by 0.62\% to 1.50\% on CIFAR-10 dataset and 0.38\% to 1.71\% on CIFAR-100 dataset. These results 
suggest that the on-line soft labels are more reliable due to the better decoupling between the past average models of the two networks, which can effectively mitigate the negative influence of hard noisy labels and avoid bias amplification even when the networks have much erroneous outputs in the early training epochs.

\subsection{Effectiveness of different transforms} We conduct experiments to investigate that the prediction consistency under different image transforms on CIFAR-10 and CIFAR-100 datasets. Specifically, we focus on a subset of frequently used transforms: \emph{scaling, rotation}, and \emph{ flipping}. Scaling means the input images are resized to 48 $\times$ 48, and then cropped to 32 $\times$ 32. Rotation contains $90{}^\circ $, $180{}^\circ $, $270{}^\circ $ and $360{}^\circ $, and flipping contains vertical flipping and horizontal flipping. The results are shown in Table~\ref{transform}. Among them, the test accuracy of horizontal flipping gains the best performance on two datasets, and the test accuracy of scaling is the worst on two datasets except two cases on CIFAR-10, \ie, Symmetry-50\% and Symmetry-80\%. While for rotation and vertical flipping, the results in four cases on two datasets present comparable performance.

\section{Conclusion}
In this paper, we propose a simple and effective approach that utilize the transform consistency to identify mislabeled samples. Specifically, we train one single network with a joint loss between two inputs (the original image and its transformed image), which includes two classification losses and one KL loss. Furthermore, we design a classification loss by using the off-line hard labels and on-line soft labels to provide more reliable supervisions for training a robust model. We conduct comprehensive experiments on CIFAR-10, CIFAR-100 and Clothing1M datasets and achieve the state-of-the-art performance.

\bibliographystyle{ieeetr}
\bibliography{egbib}
\end{document}